\documentclass{article}
\PassOptionsToPackage{numbers, compress}{natbib}
\usepackage[preprint]{neurips_2025}
\usepackage[utf8]{inputenc} 
\usepackage[T1]{fontenc}    
\usepackage{hyperref}       
\usepackage{url}            
\usepackage{booktabs}       
\usepackage{amsfonts}       
\usepackage{nicefrac}       
\usepackage{microtype}      
\usepackage{xcolor}         
\usepackage{graphicx}
\usepackage{siunitx}
\usepackage{tcolorbox}
\usepackage{adjustbox}
\usepackage{tikz}
\usetikzlibrary{positioning}
\usepackage{colortbl} 
\usepackage{multirow} 
\title{ShapeLLM-Omni: A Native Multimodal LLM for 3D Generation and Understanding}
\author{
  \textbf{Junliang Ye}$^{1,3}$\footnotemark[1]\quad
  \textbf{Zhengyi Wang}$^{1,3}$\footnotemark[1]\quad  
  \textbf{Ruowen Zhao}$^{1}$\thanks{Equal contribution} \quad
  \textbf{Shenghao Xie}$^{2}$\quad
  \textbf{Jun Zhu}$^{1,3}$\thanks{Corresponding author.} \quad \\
  Tsinghua University$^{1}$\quad Peking University$^{2}$\quad ShengShu $^{3}$ \\
  \url{https://github.com/JAMESYJL/ShapeLLM-Omni/}
}

\begin{document}
\maketitle
\begin{abstract}
Recently, the powerful text-to-image capabilities of GPT-4o have led to growing appreciation for native multimodal large language models. However, its multimodal capabilities remain confined to images and text. Yet beyond images, the ability to understand and generate 3D content is equally crucial. To address this gap, we propose ShapeLLM-Omni—a native 3D large language model capable of understanding and generating 3D assets and text in any sequence. First, we train a 3D vector-quantized variational autoencoder (VQVAE), which maps 3D objects into a discrete latent space to achieve efficient and accurate shape representation and reconstruction. Building upon the 3D-aware discrete tokens, we innovatively construct a large-scale continuous training dataset named 3D-Alpaca, encompassing generation, comprehension, and editing, thus providing rich resources for future research and training. Finally, by performing instruction-based training of the Qwen-2.5-vl-7B-Instruct model on the 3D-Alpaca dataset. Our work provides an effective attempt at extending multimodal models with basic 3D capabilities, which contributes to future research in 3D-native AI.
\end{abstract}
\section{Introduction}
\begin{figure}[ht]
    \centering
    \includegraphics[width=\linewidth]{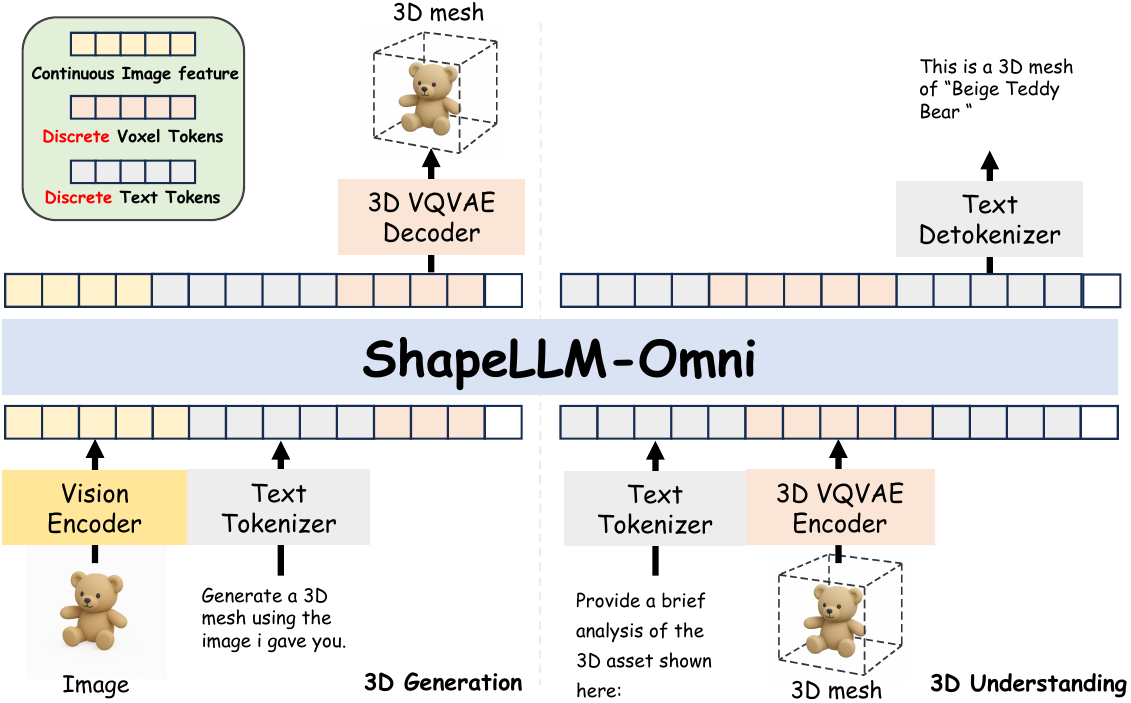}
    \vspace{-0.03\textheight}
    \caption{ShapeLLM-Omni inherits Qwen2.5-vl’s strong multimodal capabilities and additionally supports text-to-3D, image-to-3D, 3D captioning, and 3D editing using text instruction.}
    \label{fig:pipeline}
\end{figure}

Large language models have made significant achievements, including text-only language models (LLMs)~\cite{achiam2023gpt,liu2024deepseek,bai2023qwen,touvron2023llama}, Multimodal Large Language Models (MLLMs) that can understand images~\cite{hurst2024gpt,glm2024chatglm,Chameleon_Team_Chameleon_Mixed-Modal_Early-Fusion_2024}, video~\cite{guo2025fila,cheng2024videollama,maaz2023video,li2024llama} and 3D~\cite{wang2024llama,siddiqui2024meshgpt,chen2023pointgpt,chen2024sar3d} content. These models employ similar transformer architectures, using dedicated encoders to model each modality independently, thereby integrating images, video, and 3D modalities into existing LLMs.

Recently, ChatGPT-4o~\cite{hurst2024gpt} has demonstrated remarkable performance. By natively incorporating image generation and understanding into the large language model (LLM) architecture, it enables more fine-grained and precise control through human instructions. However, its multimodal capabilities remain confined to images and text, limiting its potential in more complex spatial domains. 

In this work, we propose a unified approach to integrate 3D generation and understanding into a pre-trained multimodal large language model (MLLM). Enhancing LLMs with native 3D capabilities is crucial for downstream applications such as 3D content creation, robotics, digital twins, and immersive virtual environments.

Our method adopts a fully next-token prediction paradigm, which ensures natural compatibility with joint training and large-scale scalability. We leverage a VQVAE to encode 3D meshes into compact discrete tokens, enabling a unified representation. These tokens are utilized for both understanding and generating 3D meshes, following a format analogous to language modeling.

To enable LLMs with 3D ability, we construct a comprehensive training dataset using 3D shapes from a mixture of 3D datasets~\cite{deitke2023objaverse,deitke2023objaversexl,collins2022abo,chang2015shapenet}. We construct interleaved 710k text/image-3D pairs to enable the model for basic 3D understanding ability and text/image to 3D generation ability.

Furthermore, to enable interactive 3D mesh editing, we introduce a novel dataset of 62k paired 3D meshes and corresponding text-based editing instructions. This facilitates fine-grained manipulation of 3D assets through natural language, making real-time editing more intuitive and controllable.

After that, we train an LLM on the corpus. We resume from Qwen-2.5-VL-Instruct-7B~\cite{bai2025qwen25vl} to utilize the effective of its large-scale pre-training on text and images. Our model demonstrates a wide range of capabilities, including: (1) generating 3D content from language instructions; (2) generating 3D objects from image inputs; (3) interactively editing 3D assets using natural language; (4) understanding and interpreting 3D meshes for semantic and geometric reasoning.

In all, our contributions are:
\begin{itemize}
    \item We propose a novel framework for unified 3D object generation and understanding based on a fully autoregressive next-token prediction paradigm.
    \item We present the 3D-Alpaca dataset for training large language models (LLMs) with 3D capabilities. Comprising 3.46 billion tokens, it covers three core tasks: 3D generation, 3D understanding, and 3D editing.
    \item Our experimental results provide strong empirical evidence supporting the effectiveness of the proposed method.
\end{itemize}
\section{Related Work}

\subsection{3D Mesh Generation}

The remarkable achievement of 2D diffusion models~\cite{ho2020denoising, rombach2022high} has facilitated the exploration of 3D generative models. Early 3D generation methods ~\cite{poole2022dreamfusion,wang2023prolificdreamer,chen2023fantasia3d,lin2023magic3d,raj2023dreambooth3d,sweetdreamer,sun2023dreamcraft3d,chen2024text,sjc,tang2023dreamgaussian,yi2024gaussiandreamer} often rely on SDS-based optimization to distill 3D content due to the limited 3D data, but encounter challenges such as long optimization time and Janus problem. Subsequent works such as ~\cite{wang2023imagedream,shi2023mvdream,wang2023animatabledreamer,ye2024dreamreward,qiu2024richdreamer,chen2024microdreamer} enhance semantic consistency across different views during multi-view image synthesis. To minimize generation time, more recent approaches~\cite{long2024wonder3d,zhao2024flexidreamer,liu2023syncdreamer,liu2023zero,shi2023zero123++,weng2023consistent123,liu2023one,wu2024unique3d,chen2024v3d,voleti2024sv3d,ye2024stablenormal,liu2024reconxreconstructscenesparse} adopt a two-stage pipeline that integrates multi-view image prediction with 3D reconstruction to produce 3D models. LRM ~\cite{hong2023lrm} and other works ~\cite{tang2024lgm,wei2024meshlrm,ziwen2024long,li2023instant3d,xu2023dmv3d,wang2023pf,NEURIPS2024_123cfe7d,zhang2024geolrm,zhang2024gs,zou2024triplane,xu2024instantmesh,nawrot2021hierarchical,wang2024crm} build on a feed-forward reconstruction model and predict 3D structures within seconds. Additionally, native 3D diffusion models ~\cite{zhao2023michelangelo,wang2023rodin,wu2024direct3d,yang2024hunyuan3d,huang2025spar3d,yang2024hunyuan3d,zhang2024clay,xiang2024structured,chen20243dtopia,li2024craftsman,wu2024direct3d,ye2025hi3dgen} encode 3D objects into a VAE latent and adapt a latent diffusion model on the resulting representations for comprehensive 3D understanding. Nevertheless, the above methods treat 3D objects as numerical fields~\cite{mildenhall2021nerf,kerbl20233d} and extract meshes using Marching Cubes~\cite{lorensen1998marching}, which are not easily represented as discrete tokens.

\subsection{Autoregressive 3D Generation}

Inspired by the success of auto-regressive models in language and image synthesis, some pioneering works~\cite{siddiqui2024meshgpt,chen2024meshanything,weng2024pivotmesh} have explored their use in 3D shape generation. They adopt VQVAE~\cite{van2017neural} to compress 3D shapes into latent spaces, which are subsequently quantized into discrete tokens for learning via an auto-regressive transformer. Instead of employing VQVAE, other studies ~\cite{chen2024meshanythingv2,chen2025meshxl,weng2024scaling,tang2024edgerunner,hao2024meshtron,zhao2025deepmesh} have proposed specialized mesh tokenization techniques that transform mesh vertices and faces into compact discrete token sequences, while preserving the original complex geometric details. These approaches enable the auto-regressive model to effectively generate meshes in a face-by-face manner. Building on 3D auto-regressive models, LLaMA-Mesh~\cite{wang2024llama} explores the integration of natural language instructions with mesh generation and understanding, enabling interactive 3D content creation through a unified framework. However, it treats the 3D OBJ mesh file as text for language model to process, which overlooks the inherent topological structures of 3D data. 

\subsection{Unified Models for Multimodal Understanding and Generation}
Extending large language models (LLMs) to process, generate, and comprehend multiple modalities—such as vision and language—within a unified framework has become a major research frontier. Previous studies~\cite{bai2023qwen,chen2024internvl,alayrac2022flamingo} have advanced this direction by equipping LLMs with visual understanding capabilities for multimodal tasks. Concurrently, other works~\cite{Chameleon_Team_Chameleon_Mixed-Modal_Early-Fusion_2024,liu2024world,wang2024emu3,xie2024show,zhou2024transfusion} have proposed the integration of image and text generation through specialized visual tokenizers. More recently, ChatGPT-4o has further propelled this progress, achieving state-of-the-art performance in both visual comprehension and image synthesis.
Beyond 2D modalities, a growing body of research~\cite{hong20233d,xu2024pointllm,qi2024shapellm,xue2023ulip} has extended LLMs to 3D content understanding, primarily through point cloud representations. However, point clouds often lack fine-grained geometric detail and are challenging to acquire in real-world settings, limiting their applicability for interactive generation. Despite these advancements, there remains a notable gap: very few models are capable of jointly processing and generating text, images, and 3D data in an integrated manner.
To bridge this gap, we introduce a 3D VQVAE module that encodes 3D shapes into discrete representations, enabling autoregressive models to perform unified multimodal understanding and generation across text, images, and 3D content.

\begin{figure}[t]
    \centering
    \includegraphics[width=\linewidth]{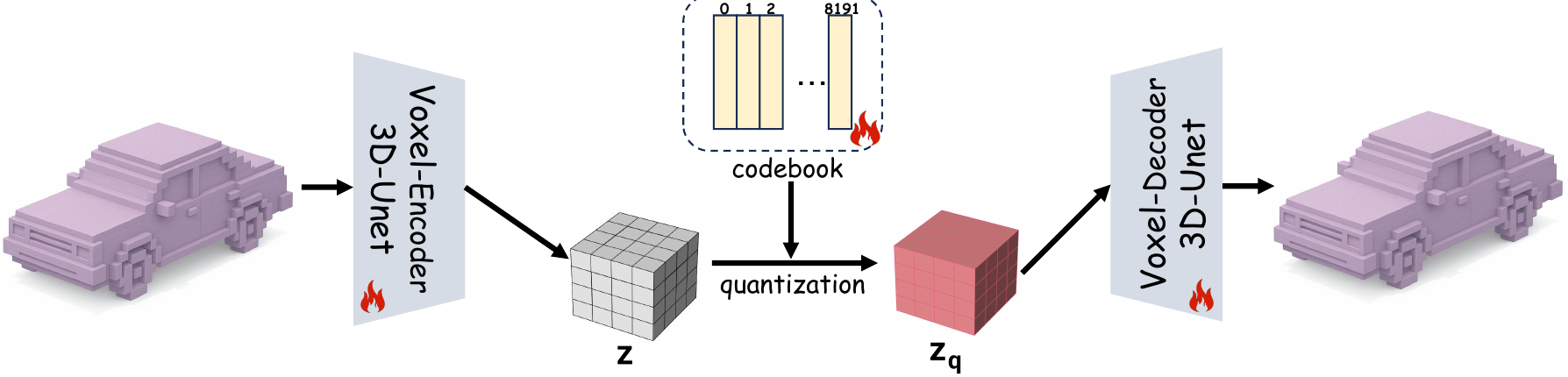}
    \vspace{-0.03\textheight}
    \caption{The pipeline of 3D VQVAE, which can compress voxels into discrete tokens.}
    \label{fig:vqvae_pipeline}
\end{figure}

\section{Method}
\begin{table}[ht]
  \centering
  \caption{\textbf{Modality comparison}. In contrast to the task-specific model architectures of SAR3D and Trellis, ShapeLLM-Omni achieves cross-modal alignment by jointly modeling text and 3D representations in a shared latent space, enabling unified understanding and generation capabilities.}
  \begin{tabular}{l|>{\columncolor{cyan!20}}c>{\columncolor{purple!20}}c>{\columncolor{blue!20}}c>{\columncolor{gray!20}}c|>{\columncolor{cyan!20}}c>{\columncolor{purple!20}}c>{\columncolor{blue!20}}c}
    \toprule
    \multirow{2}{*}{} 
      & \multicolumn{4}{c|}{Input Modality} 
      & \multicolumn{3}{c}{Output Modality} \\
    \cline{2-8}
      & Text & Image & 3D & Unified model 
      & Text & Image & 3D \\
    \hline
    SAR3D~\cite{chen2024sar3d}     &\checkmark&\checkmark&\checkmark&&\checkmark&      &\checkmark \\
    \hline
    Trellis~\cite{xiang2024structured}   &\checkmark&\checkmark&&&&&\checkmark    \\
    \hline
    PointLLM~\cite{xu2024pointllm}  &\checkmark&&\checkmark&\checkmark&\checkmark&&\\
    \hline
    LLaMA-Mesh~\cite{wang2024llama}&\checkmark&&\checkmark&\checkmark&\checkmark&&\checkmark\\
    \hline
    ChatGPT-4o~\cite{hurst2024gpt4o}&\checkmark&\checkmark&&\checkmark&\checkmark&\checkmark&\\
    \hline
    Qwen-2.5vl~\cite{bai2025qwen25vl}&\checkmark&\checkmark& &\checkmark&\checkmark&&    \\
    \hline
    \textbf{ShapeLLM-Omni (ours)}&\checkmark&\checkmark&\checkmark&\checkmark&\checkmark&      &\checkmark \\
    \bottomrule
  \end{tabular}
  \label{tab:modality}
\end{table}
\subsection{Overview}
Figure~\ref{fig:pipeline} provides an overview of our native Multimodal LLM framework, which can handle mixed sequences of text, images, and 3D data and produce corresponding text or 3D outputs. We begin by converting 3D assets into discrete tokens using a 3D VQVAE (Sec.~\ref{sec:3d-vqvae}), which allows us to leverage the same transformer architecture for both 3D and text token sequences. Subsequently, we assemble a comprehensive 3D supervised fine-tuning dataset, 3D-Alpaca (Sec.~\ref{sec:3d-alpaca}), covering text-to-3D generation, image-to-3D generation, 3D captioning, and 3D editing.
\subsection{Architecture}
As shown in Figure~\ref{fig:pipeline}, we represent both text and 3D data as sequences of discrete tokens, enabling fully autoregressive multi-modal generation. This design allows for flexible input and output across modalities in any order. While we adopt token-based representations for both text and 3D modalities, we use continuous features for images. This is because images are only involved in understanding tasks, whereas 3D data supports both understanding and generation. Such a unified modeling approach—based on early fusion—facilitates better modality integration within the language model. Compared to prior work in the 3D domain Table~\ref{tab:modality}, our model is the first unified auto-regressive framework that supports text-to-3D, image-to-3D, 3D understanding, and 3D editing in a single system. It also marks the first attempt at a ChatGPT-4o-style model tailored for 3D tasks.

\subsection{3D VQVAE}
\label{sec:3d-vqvae}
In this section, we introduce our 3D representation—voxels—explain why we chose voxels, and how we compress voxels into discrete tokens using a 3D VQVAE. Finally, we describe how to reconstruct high-quality 3D meshes from voxels.
\paragraph{Voxel-Based Representation}
3D assets can be represented in various ways—such as voxels, point clouds, or Gaussian splats~\cite{kerbl20233d}. In this work, we adopt low-resolution voxels as our representation: on one hand, they compress complex 3D information into a much smaller space, which facilitates subsequent training; on the other hand, voxels effectively preserve an asset’s essential shape and skeletal structure, providing sufficient information for a language model. Moreover, we can leverage open-source models to reconstruct coarse-resolution voxels into high-quality, detail-rich meshes.
\paragraph{Model Architecture}
We adopt a $64^3$ voxel grid resolution, as voxels at this resolution strike the optimal balance for modeling 3D skeletons, preserving essential structural details while avoiding excessive redundancy~\cite{xiang2024structured}. Although voxel representations are compact, even modeling a single 3D object with a $64^3$ voxel grid still requires $64^3$ tokens—far beyond what a large language model can handle. Therefore, we further compress voxels using a 3D VQVAE~\cite{xiang2024structured}: first, we encode the $64^3$ grid into a $16^3$ latent grid; then we serialize it into 4096 tokens. However, 4096 tokens remain too long. Inspired by \cite{Chameleon_Team_Chameleon_Mixed-Modal_Early-Fusion_2024}, which represents images as 1024 tokens, we concatenate every four neighboring tokens along the channel dimension—transforming the original 4096 tokens with 8 channels into 1024 tokens with 32 channels. Finally, we employ an 8192-entry codebook to compress the voxels into 1024 discrete tokens. In all, we represent a single 3D object using 1024 discrete tokens, for both generation and understanding.
\paragraph{Shape Reconstruction}
Although we employ voxel-based representations for 3D shape generation, practical deployment often necessitates converting voxels into meshes for downstream applications. To address this, we adopt the approach proposed by Xiang et al.~\cite{xiang2024structured}, which utilizes a Rectified Flow model to refine and complete voxel information, enabling high-quality mesh reconstruction. By first generating 3D shapes in the voxel domain and then converting them into meshes using this method, our framework achieves a balance between precision and efficiency. This hybrid representation allows large language models to exert fine-grained control over 3D content generation while avoiding the computational burden associated with high-resolution geometry.

\begin{figure}[t]
    \centering
    \includegraphics[width=\linewidth]{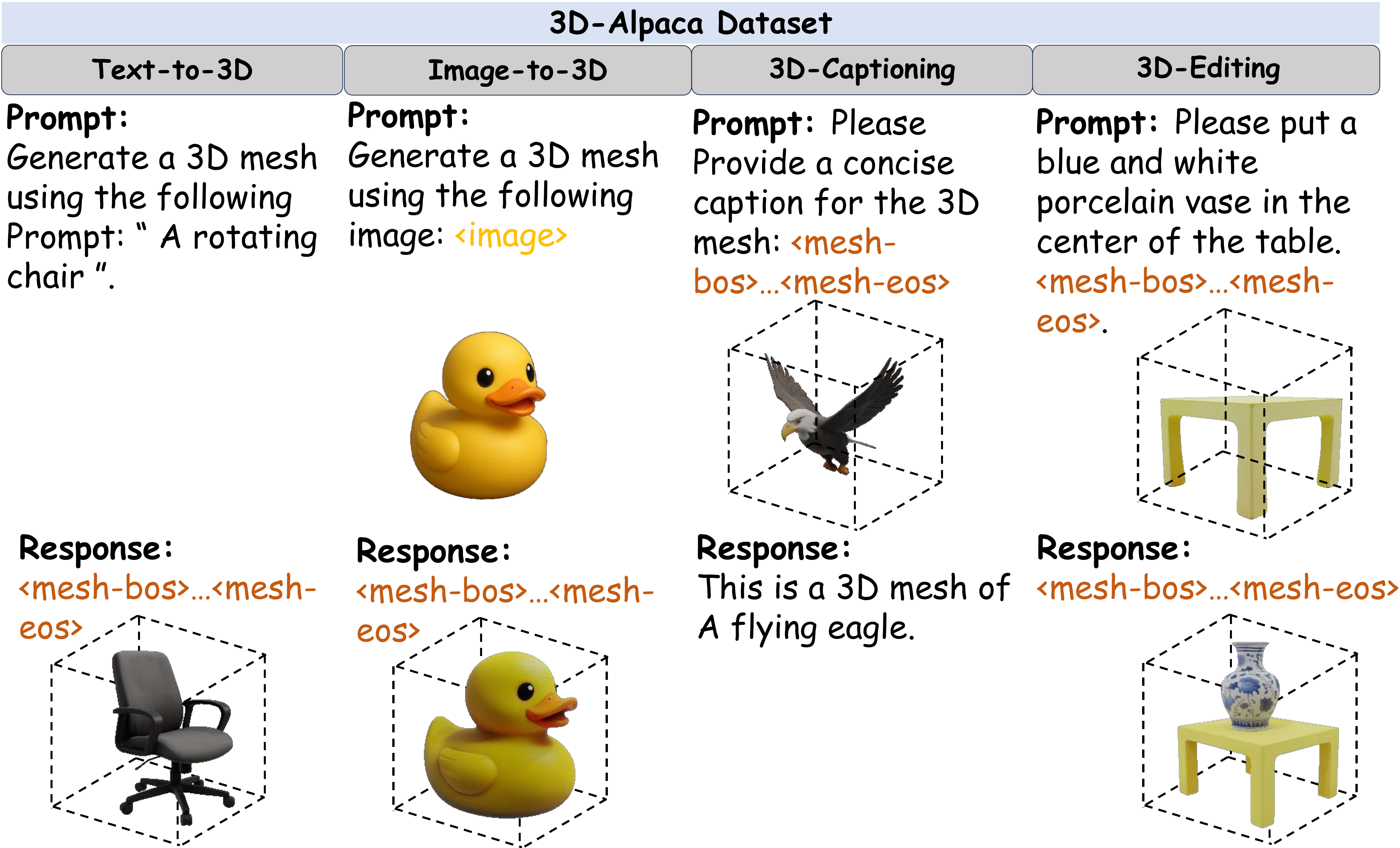}
    \vspace{-0.03\textheight}
    \caption{Our proposed 3D-Alpaca dataset comprises 3D generation, 3D understanding, and 3D editing components, providing a comprehensive foundation for training and evaluating 3D large language models.}
    \label{fig:3d-dataset}
    \vspace{-0.3cm}
\end{figure}
\subsection{3D-Alpaca Dataset Construction}
\label{sec:3d-alpaca}

Although a wealth of datasets has been developed for the supervised fine-tuning of multimodal large-language models, dialogue data within the 3D LLM~\cite{hong20233d,chen2024sar3d,xu2024pointllm} domain remains relatively scarce. To bridge this gap, we introduce 3D-alpaca, a comprehensive dataset encompassing tasks in 3D content generation, comprehension, and editing.

\paragraph{3D Generation and Understanding Dataset} 
We select a high-quality subset of approximately 712k 3D assets from Trellis~\cite{xiang2024structured} and internal collection. For the image collection, each asset is rendered into a 2D image, and a random offset is applied to the frontal view to create the input. Moreover, these rendered images also underpin the construction of the editing dataset in the Sec.~\ref{sec:3d-alpaca-edit}. To generate the text collection and enable early fusion across all three modalities, we render four orthogonal views—front, back, left, and right—of each asset. These multi-view images are then input into the base model Qwen-2.5-VL-Instruct~\cite{bai2025qwen25vl} to generate descriptive captions. The resulting captions are utilized both as prompts for text-to-3D generation and as ground-truth targets for 3D-to-text captioning tasks.
\paragraph{3D Edited Dataset}
\label{sec:3d-alpaca-edit}
We aim to build a 3D asset-editing dataset composed of paired 3D assets, where each pair is linked to a specific editing instruction. Despite recent advances in 3D content creation, the field still lacks a model capable of performing consistent edits on 3D assets. In light of the promising performance of current image-editing models, we therefore adopt an image-mediated pipeline: first rendering each 3D asset into images and applying an image-editing model, then reconstructing the edited images back into 3D assets via an image-to-3D generation method. Based on the multimodal alignment demonstrated and with the aim of equipping the model with ChatGPT-4o–level editing capabilities, we follow a six-step pipeline. 

(1) \textit{Category}: We reference the data distribution of Objaverse-XL~\cite{deitke2023objaverse} and manually selected the 100 most representative and frequent object categories, such as cars, tables, cabinets, human figures, etc. 

(2) \textit{Asset Classification}: Using ChatGPT-4o, we classify the 3D assets in our dataset into fine-grained subcategories, with the frontal view renderings of each asset as input. From the 3D asset dataset, we filtered 311k assets belonging to the predefined 100 major categories.

(3) \textit{Editing-Prompt Definition}: We provide the category names to ChatGPT-4o and instruct it to generate 20 feasible editing-prompts for each category. The instruction given to ChatGPT-4o is: "For each given category name, suggest potential image editing operations that could be applied to objects of that category." Next, we manually review each generated editing prompt and retain only those that meet both our technical feasibility and visual engagement criteria, resulting in 371 unique editing prompts (e.g: “Replace the chair’s backrest with a mesh frame”).

(4) \textit{Asset Sampling \& Annotation}: Due to time and resource constraints, we build a compact, high-quality dataset of editing prompts rather than applying every possible editing prompt to each asset. Specifically, we allocate 200 assets to each editing prompt.

(5) \textit{Editing-Image Pair Collection}: For each sampled asset, we provide ChatGPT-4o with its frontal render plus the chosen editing-prompt, and ChatGPT-4o produces the corresponding edited image, yielding image-level editing pairs. After filtering out erroneous cases, we end up with 70k valid editing samples.

(6) \textit{3D reconstruction}: Finally, we employ Trellis~\cite{xiang2024structured} to convert the curated images into 3D assets, resulting in 3D pairs before/after editing.
\paragraph{Dialogue Data Construction}
We define 25 dialogue templates per task (e.g., “Generate a 3D asset of prompt/images”) and encode all 3D assets into discrete token sequences with our pre-trained 3D VQVAE (Sec.~\ref{sec:3d-vqvae}). For each 3D-edit instance, we randomly select 6 templates from a pool of 25; for all other instances, we randomly assign one template each. By merging the tokens with these templates, we create a training corpus of 2.5 million 3D dialogues.
\paragraph{General Conversation}
To ensure the model’s general conversational capability, we adopt UltraChat~\cite{ding2023enhancing} as our text-only dataset, with its data distribution shown in the Table~\ref{tab:example}. For additional details, please refer to the Appendix.

\paragraph{Putting these together}
After data processing and construction, we finally arrive at the 3D-Alpaca dataset. As shown in the Table~\ref{tab:example}, the dataset includes four types of tasks: image-to-3D, text-to-3D, 3D-to-caption, and 3D-editing. Together, these four subsets form a total of 2.56 million samples, comprising 3.46 billion tokens. To ensure the large language model retains its original reasoning and dialogue capabilities, we additionally include the UltraChat~\cite{ding2023enhancing} dataset, a high-quality, large-scale multi-turn dialogue corpus.

\begin{table}[ht]
  \centering
  \caption{\textbf{Corpus Data Proportions} An overview of token and item counts in the training corpus, covering two datasets: the \textit{3D-Alpaca} dataset, which includes four task types—Text-to-3D, Image-to-3D, 3D-to-Caption, and 3D-Editing—and the text-only \textit{UltraChat} dataset~\cite{ding2023enhancing}}
  \begin{tabular}{c|ccccc|c}
    \toprule
     & Text-To-3D & Image-To-3D & 3D-to-Caption & 3D-Edit &3D-All&Text-Only\\ 
     \midrule
    Token count &$0.77$B  & $1.01$B &  $0.77$B &$0.91$B& $3.46$B&$2.16$B  \\ \hline
    Item count & $712$k & $712$k& $712$k& $420$k&$2.56$M&$1.47$M\\ \bottomrule
  \end{tabular}
  \label{tab:example}
\end{table}
\section{Experiments}
\subsection{Implementation Details}
For training our 3D VQVAE, we adopt a 3D U-Net VAE architecture introduced in Trellis~\cite{xiang2024structured}. Our training follows a two-stage strategy: In Stage 1, we freeze the VAE’s pre-trained parameters and train only the codebook. In Stage 2, we unfreeze the VAE and jointly fine-tune it with the codebook. Concretely, each stage runs for 1000 steps on 48 NVIDIA H100 GPUs with a batch size of 25, while the learning rate decays from $5\times 10^{-3}$ to $5\times 10^{-5}$. For the training of ShapeLLM-Omni, we use Qwen-2.5-VL-Instruct-7B~\cite{bai2025qwen25vl}, a multimodal large language model (MLLM) with image-understanding capability, as our backbone. Specifically, we extend its base architecture by adding the 8192 3D VQVAE codebook. To preserve its original image-understanding skills, we freeze the parameters of Qwen2.5-vl’s visual encoder. While training, the learning rate decays from $5\times 10^{-5}$ to $5\times10^{-6}$, with a per-GPU batch size of 2 and gradient accumulation over 2 steps. The model is trained for 15 epochs on 48 NVIDIA H100 GPUs.
\subsection{Quantitative comparisons}
\paragraph{Language and Conversational Abilities}
Table~\ref{tab:language-results} presents quantitative results evaluating language abilities. The table provides a comparison with models: LLaMA-Mesh~\cite{wang2024llama}, Chameleon~\cite{Chameleon_Team_Chameleon_Mixed-Modal_Early-Fusion_2024}, and Qwen2.5-vl~\cite{bai2025qwen25vl}. The metrics include SIQA~\cite{sap2019socialiqa}, PIQA~\cite{bisk2020piqa}, MMLU~\cite{hendrycks2020measuring}, and GSM8K~\cite{cobbe2021training}. Fine-tuned on 3D-Alpaca for both 3D mesh generation and comprehension, our ShapeLLM-Omni maintains language understanding and reasoning performance on par with baseline models. The result demonstrates that ShapeLLM-Omni effectively extends the MLLM’s capabilities to 3D content generation while preserving its native language capabilities.

\begin{table}[ht]
  \centering
  \caption{\textbf{Language capabilities comparison}. We provide a comparison with models: LLaMA-Mesh~\cite{wang2024llama}, Chameleon~\cite{Chameleon_Team_Chameleon_Mixed-Modal_Early-Fusion_2024}, and Qwen2.5-vl~\cite{bai2025qwen25vl}. The metrics include SIQA~\cite{sap2019socialiqa}, PIQA~\cite{bisk2020piqa}, MMLU~\cite{hendrycks2020measuring}, and GSM8K~\cite{cobbe2021training}. Fine-tuned on 3D-Alpaca for both 3D mesh generation and comprehension, our ShapeLLM-Omni maintains language understanding and reasoning performance. The table highlights the optimal values in bold and the suboptimal values with underlining.}
  
  \label{tab:language-results}
  \begin{tabular}{c|cccc}
    \toprule
    Metric   & Qwen2.5-vl-7B &ShapeLLM-Omni-7B & Chameleon-7B & LLaMA-Mesh-8B \\
    \midrule
    MMLU      & \textbf{66.9}         & \underline{63.9}           & 59.4         & 57.4         \\
    PIQA      & \textbf{81.0}         & 78.6           & \underline{79.6}         & 78.9         \\
    GSM8K     & 42.9         & \underline{55.1}           & \textbf{66.9}         & 33.1         \\
    SIQA      & 40.7         & \underline{41.0}           &           \textbf{57}          & 40.4         \\
    \bottomrule
  \end{tabular}
\end{table}

\paragraph{3D Generation}
We compare our methods on both text-to-3D and image-to-3D generation tasks against CRM~\cite{wang2024crm}, SAR3D~\cite{chen2024sar3d}, 3DTopia-XL~\cite{chen20243dtopia}, and TRELLIS~\cite{xiang2024structured}. When evaluating the generation performance of ShapeLLM-Omni, we set the model's top-k parameter equal to the size of the 3D vocabulary (8192), with top-p=0.7 and temperature=0.7. Regarding the dialogue templates, the image-to-3D template is formulated as: \textit{"Create a 3D asset using the following image: <image>"}, while the text-to-3D template is expressed as: \textit{"Please generate a 3D mesh based on the prompt I provided: <prompt>"}. Quantitative evaluations are conducted using image and text prompts sampled from the Toys4K~\cite{stojanov2021using} test dataset, with the results summarized in Table~\ref{tab:gen-metric}. To assess the overall quality of the generated 3D outputs, following~\cite{xiang2024structured}, we compute Frechet Distance (FD)~\cite{heusel2017gans} and Kernel Distance (KD)~\cite{binkowski2018demystifying} using Inception-V3~\cite{szegedy2016rethinking} features. Additionally, we report the CLIP score~\cite{radford2021learning} to measure the semantic alignment between the generated outputs and their input prompts. As shown in the Table~\ref{tab:text-image-3d}, our generation results outperform all baseline methods except for Trellis.
\paragraph{Compared with Trellis} Our results are not as good as Trellis for a few reasons. First, Trellis uses separate models for text-to-3D and image-to-3D tasks. In contrast, our ShapeLLM-Omni handles both tasks in a single model, and it also supports 3D editing, understanding, and interactive conversation. This all-in-one training comes with trade-offs and can reduce generation quality. Second, Trellis is built on a Rectified Flow model, while ours is a discrete autoregressive model. From an architectural point of view, it's expected that this may lead to some performance differences.

\begin{table}[ht]
  \centering
  \caption{Comparison of methods on Text-to-3D and Image-to-3D tasks. We scale KD by ($\times 10^2$).}
  \label{tab:text-image-3d}
  \begin{tabular}{l|ccc|ccc}
    \toprule
    \multirow{2}{*}{Method}
      & \multicolumn{3}{c|}{Text-to-3D}
      & \multicolumn{3}{c}{Image-to-3D} \\
    & CLIP$\uparrow$
    & FD$_\mathrm{incep}\downarrow$
    & KD$_\mathrm{incep}\downarrow$
    & CLIP$\uparrow$
    & FD$_\mathrm{incep}\downarrow$
    & KD$_\mathrm{incep}\downarrow$\\
    \midrule
    CRM     & - & - & - &
                 76.1 & 14.7 &  0.12 \\
    3DTopia-XL  & - & - & - & 
                    76.5 &49.5   & 1.63 \\
     SAR3D       &23.9  & 27.2 &0.28 & 
                  84.70  & 20.6  & 0.17  \\
    Trellis        & \textbf{30.8} & \textbf{18.3} & \textbf{0.19} & 
                     \textbf{85.0}& \textbf{8.31}   & \textbf{0.07} \\
    \textbf{ShapeLLM-Omni (ours)} & \underline{26.7} & \underline{25.9} &\underline{0.25}  & 
                     \underline{84.5}&\underline{12.2}   & \underline{0.09} \\
    \bottomrule
  \end{tabular}
  \label{tab:gen-metric}
\end{table}

\begin{figure}[t]
    \centering
    \includegraphics[width=\linewidth]{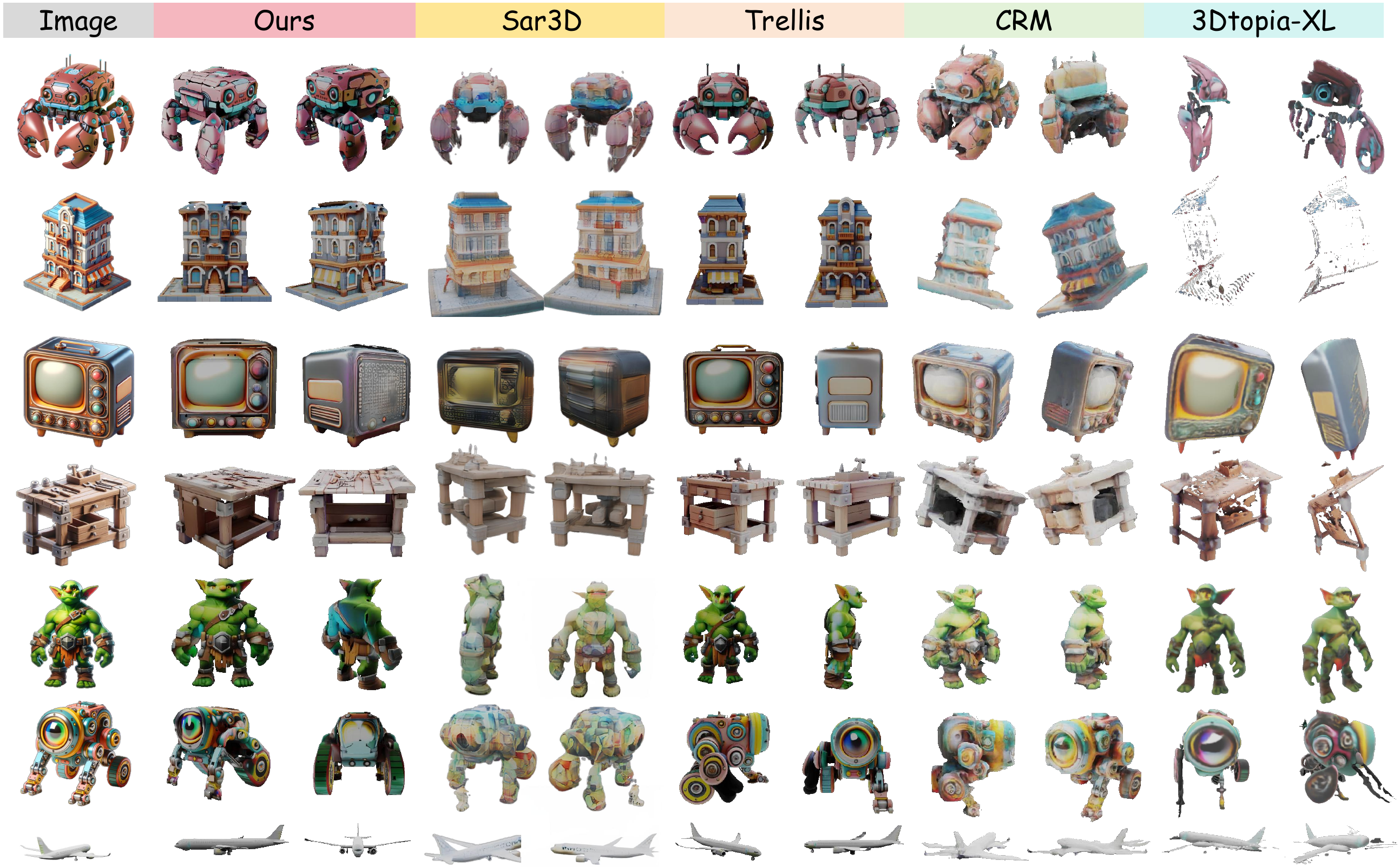}
    \caption{\textbf{Comparisons with other baselines on the image-to-3D task.} Our results demonstrate more complete geometry and high-fidelity textures compared to baselines, enabling photorealistic image-to-3D generation. }
    \label{fig:compare_image_to_3d}
\end{figure}

\paragraph{3D-to-Caption}
Following the evaluation settings provided by PointLLM~\cite{xu2024pointllm}, we test the same metrics on the benchmark dataset used by PointLLM. We adopt the same curated test set to assess the 3D-to-caption task. The dialogue prompt is structured as: \textit{“<mesh>. Caption this 3D model in detail.”}. As shown in Table~\ref{tab:combined-model-comparison}, our ShapeLLM-Omni demonstrates strong 3D understanding capabilities, with performance second only to PointLLM, which is specifically tailored for single-task 3D understanding.
\begin{table}[ht]
  \centering
\caption{\textbf{3D object captioning results~\cite{xu2024pointllm} on Objaverse~\cite{deitke2023objaverse}}. As can be seen from the table, our model achieves better performance on 3D understanding/caption tasks. "*" indicate PointLLM was prompted for shorter captions with no more than 20 words.}
  \label{tab:combined-model-comparison}
  \begin{tabular}{l|ccc|cc}
    \toprule
    Model & BLUE-1 & ROUGE-L & METEOR & Sentence-BERT & SimCSE \\
    \midrule
    InstructBLIP-13B~\cite{wenliang2023instructblip}
      &  4.65  &  8.85  & 13.23  & 45.90 & 48.86 \\
    LLaVA-13B~\cite{liu2023visual}
      &  4.02  &  8.15  & 12.58  & 46.37 & 45.90 \\
    3D-LLM~\cite{hong20233d}
      & 16.91 & 19.48 & 19.73  & 44.48 & 43.68 \\
    PointLLM-13B~\cite{xu2024pointllm}
      &  3.38  &  7.23  & 12.26  & 47.91 & 49.12 \\
    PointLLM-13B*~\cite{xu2024pointllm}
      & 17.09 & 20.99 & 16.45  & \textbf{50.15} & \textbf{50.83} \\
    \midrule
    \textbf{ShapeLLM-Omni (ours)}
      & \textbf{18.51} & \textbf{21.37} & \textbf{19.89}  & \underline{48.34} & \underline{49.72} \\
    \bottomrule
  \end{tabular}
\end{table}

\subsection{Qualitative comparisons}
\begin{figure}[t]
    \centering
    \includegraphics[width=\linewidth]{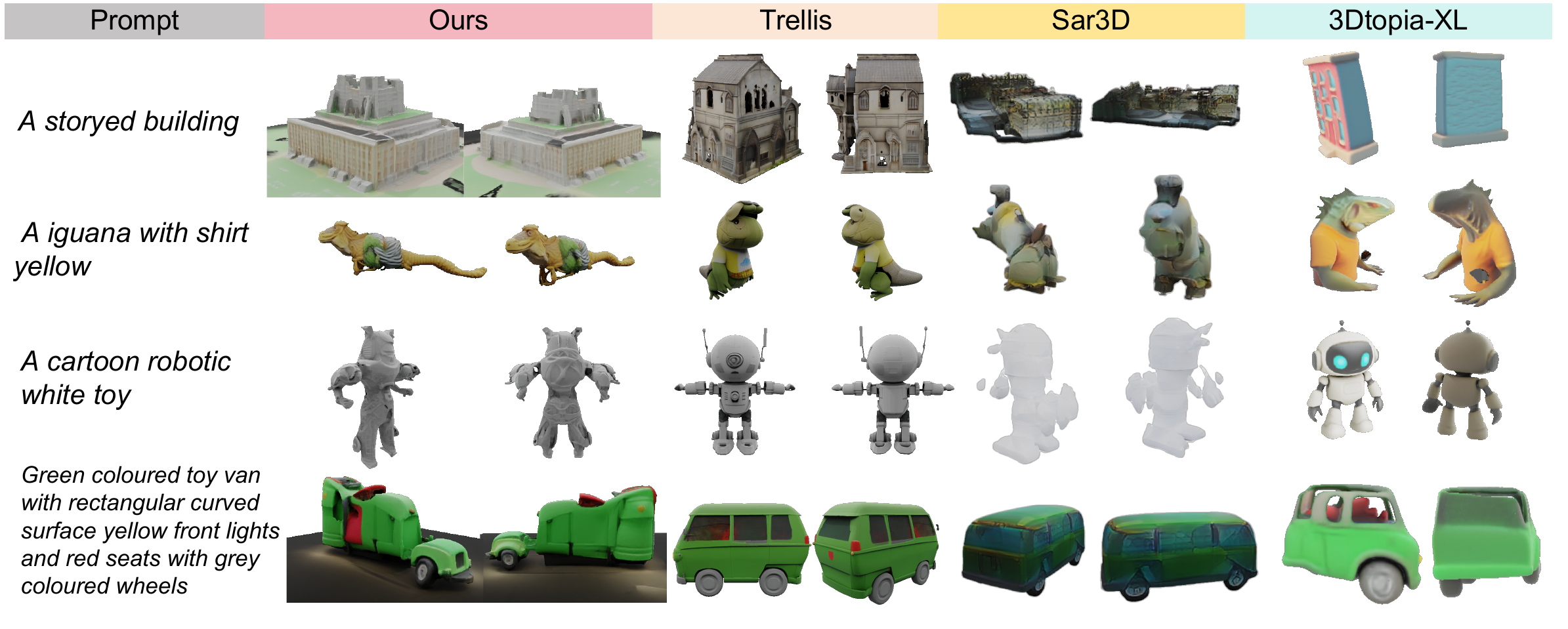}
    \caption{\textbf{Comparisons with other baselines on text-to-3d task.} Compared to other methods, our results show better text alignment, with generated 3D shapes accurately reflecting the input descriptions.}
    \label{fig:compare_text_to_3d}
\end{figure}

\paragraph{3D Generation}
To evaluate the effectiveness of our image-conditioned generation, we compare against baselines including SAR3D, TRELLIS, CRM, and 3Dtopia-XL. As illustrated in Figure~\ref{fig:compare_image_to_3d}, the baselines exhibit notable limitations in capturing fine-grained visual features, suffering from geometric distortions and texture misalignments. In contrast, our method generates high-quality 3D meshes that accurately preserve both geometry and appearance details. Moreover, our generation quality matches that of TRELLIS, which is our base model and performance upper bound, due to the integration of a well-trained 3D VQVAE and a carefully constructed 3D image-to-3d dataset for LLM fine-tuning. For text-to-3D tasks, Figure~\ref{fig:compare_text_to_3d} presents qualitative comparisons among different baselines. The input prompts are randomly generated by ChatGPT-4o to cover a diverse range of objects. Since 3Dtopia-XL does not support text-to-3D tasks, we use ChatGPT-4o to generate reference images from the test prompts. These images are then used as input for image-to-3D generation. It is evident that our method achieves precise alignment with the given text prompts and excels at generating intricate, coherent details.

\begin{figure}[th]
    \centering
    \includegraphics[width=\linewidth]{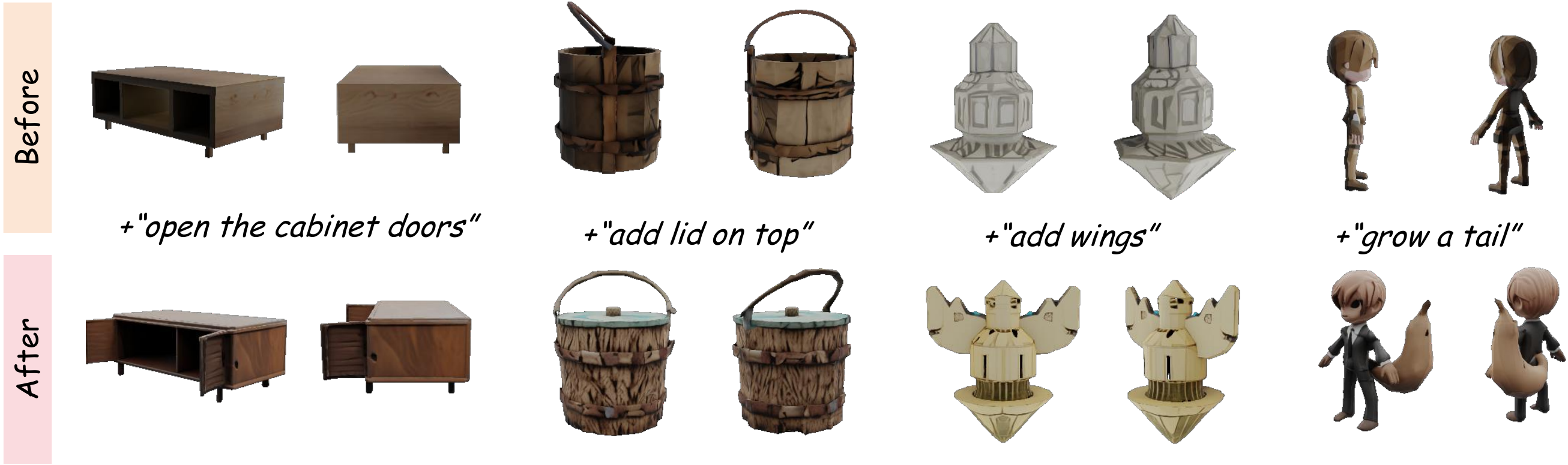}
    \caption{\textbf{Some cases of 3D editing result from our method.} Our method enables the editing of 3D assets based on textual instructions while preserving their original identity and visual consistency.}
    \label{fig:edit}
\end{figure}

\paragraph{3D Editing}
Compared with traditional generative models, native multimodal LLMs not only enhance image understanding capabilities, but also exhibit significantly improved comprehension of text instructions. Therefore, it introduces a more powerful language-driven interactive 3D asset manipulation paradigm for artists, offering a more flexible and accessible alternative to conventional software-based 3D content creation pipelines, which are usually time-consuming. As shown in Figure~\ref{fig:edit}, ShapeLLM-Omni can edit 3D assets according to user-provided instructions while maintaining good identity consistency.

\section{Conclusion}
In this work, we introduce ShapeLLM-Omni, a novel framework that advances both 3D generation and understanding through a 3D VQVAE. By constructing a comprehensive 3D-Alpaca dataset, we provide a data foundation to support future research on native 3D-modality large language models.

\textbf{Limitation} Constrained by limited resources, we possess only 70k 3D-editing pairs—far too few to achieve ChatGPT-4o–level results in 3D editing. Due to limited computing resources, our ShapeLLM-Omni only has 7B parameters. As a result, our performance hasn't yet reached the level of a true “3D version of ChatGPT-4o”.

\appendix
\section{More details about 3D-Alpaca}
\subsection{3D Editing Prompt List}
As shown in Table~\ref{table:edited1} and Table~\ref{table:edited2}, we present 70 out of the 100 categories from the 3D editing dataset, along with their corresponding editing prompts.
\subsection{3D Editing Data}
\begin{figure}[th]
    \centering
    \includegraphics[width=\linewidth]{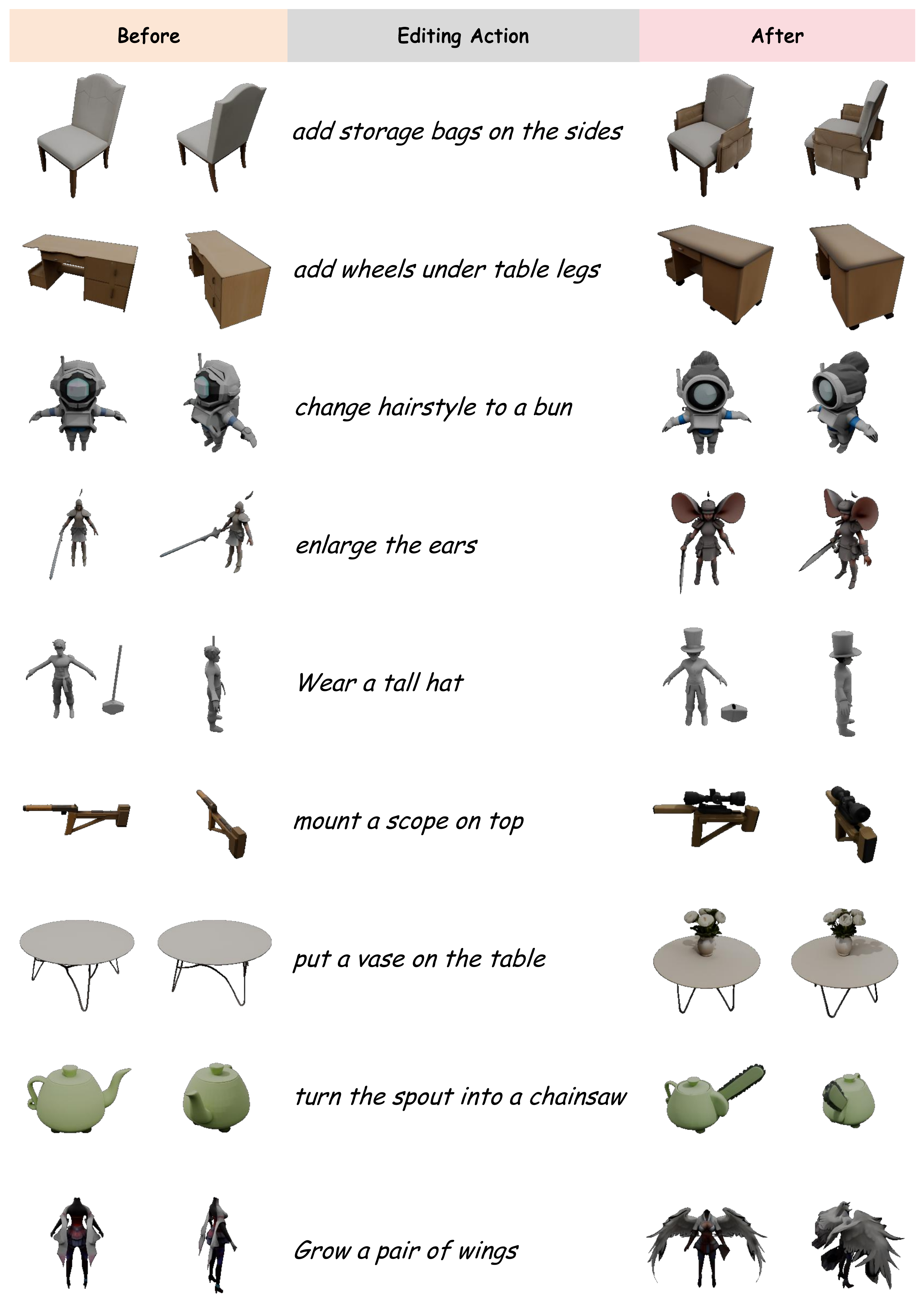}
    \caption{\textbf{Some cases of our 3D-Editing Data}}
    \label{fig:edit_data_case}
\end{figure}

As shown in Figure~\ref{fig:edit_data_case}, we present several examples from our 3D editing dataset. The figure illustrates that our 3D editing data pairs support effective modifications while preserving subject consistency between the original and edited versions.
\section{More Experiments}
\subsection{More Implementation details}
\begin{figure}[th]
    \centering
    \includegraphics[width=\linewidth]{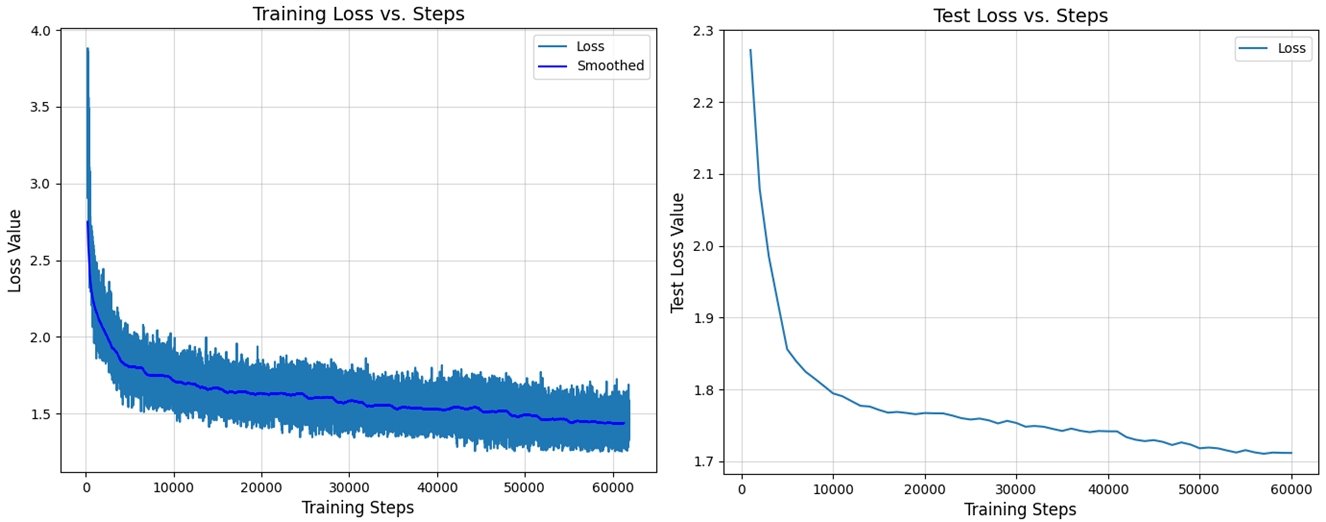}
    \vspace{-0.03\textheight}
    \caption{\textbf{Training Loss Curve and Testing Loss Curve}}
    \label{fig:loss_curve}
\end{figure}

\paragraph{Decoding Voxel into 3D Mesh}
\begin{figure}[th]
    \centering
    \includegraphics[width=\linewidth]{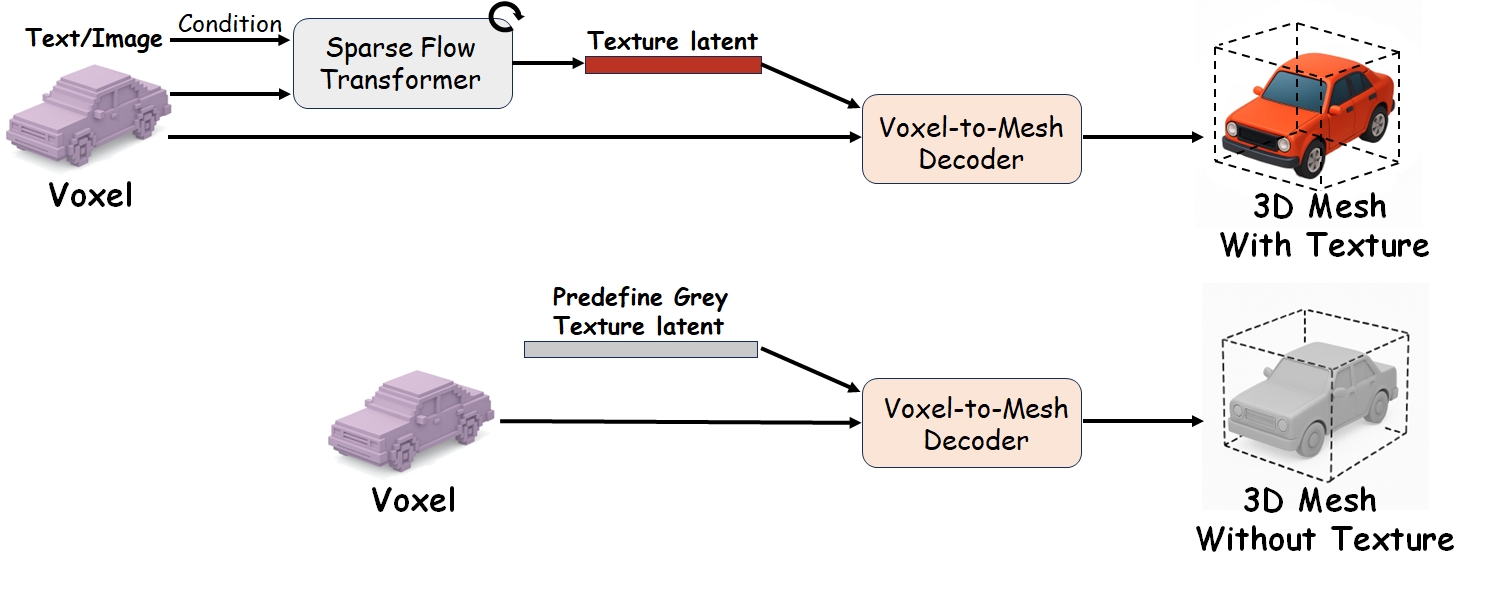}
    \vspace{-0.03\textheight}
    \caption{\textbf{About how to generate 3d mesh from voxel.} The upper part illustrates the process of reconstructing a textured mesh from voxel inputs using a texture transformer~\cite{xiang2024structured} and mesh decoder. In contrast, the lower part demonstrates the pipeline for reconstructing a non-textured mesh directly from voxels. Both reconstruction pathways are optional and can be flexibly applied based on the needs of specific applications.
}
    \label{fig:voxel_to_mesh_pipeline}
\end{figure}
As illustrated in the upper part of Figure~\ref{fig:voxel_to_mesh_pipeline}, we first utilize a texture transformer, named Sparse-Flow Transformer~\cite{xiang2024structured}, to extract texture latents from the voxel representation. These latents are then fed into a voxel-to-mesh decoder, which generates a mesh with associated texture information. Interestingly, we observe that the geometry of the output mesh is entirely determined by the input voxel representation, regardless of the presence of texture information. Inspired by this observation, and as illustrated in the lower part of Figure~\ref{fig:voxel_to_mesh_pipeline}, we define a grey texture latent to support the generation of non-textured meshes.
\paragraph{More Details about Training}
The model is trained on 48 H100 GPUs for 60k iterations. We conduct full parameter fine-tuning. We use the AdamW optimizer, with a learning rate of 1e-5, a warm-up of 400 steps with cosine scheduling, and a global batch size of 192. The total training time is around 5 days. We present the training and testing loss curve in Figure~\ref{fig:loss_curve}, which demonstrates that the model achieves rapid convergence on the new modality, reflecting effective knowledge adaptation. Throughout training, the loss remains smooth and stable, with no noticeable spikes or instability.
\subsection{More Qualitative comparisons}
\begin{figure}[t]
    \centering
    \includegraphics[width=\linewidth]{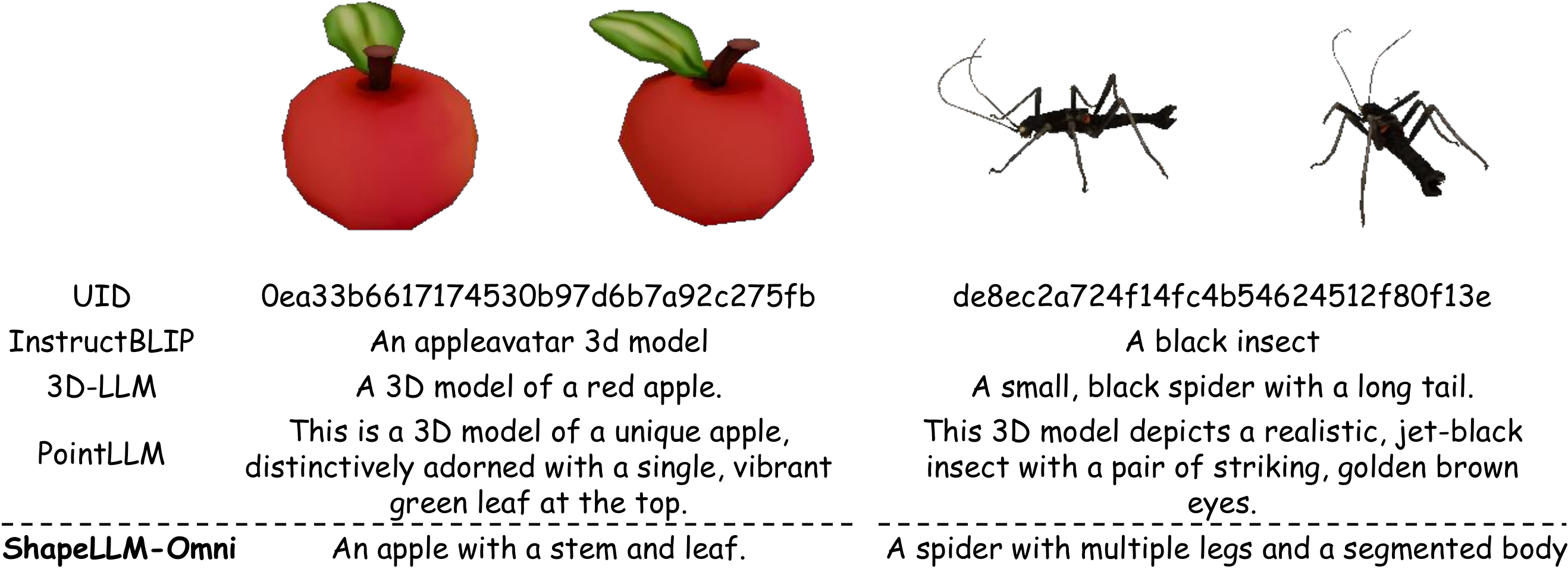}
    \vspace{-0.03\textheight}
    \caption{\textbf{Qualitative results on Objaverse.}}
    \label{fig:caption_result2}
\end{figure}
\begin{figure}[t]
    \centering
    \includegraphics[width=\linewidth]{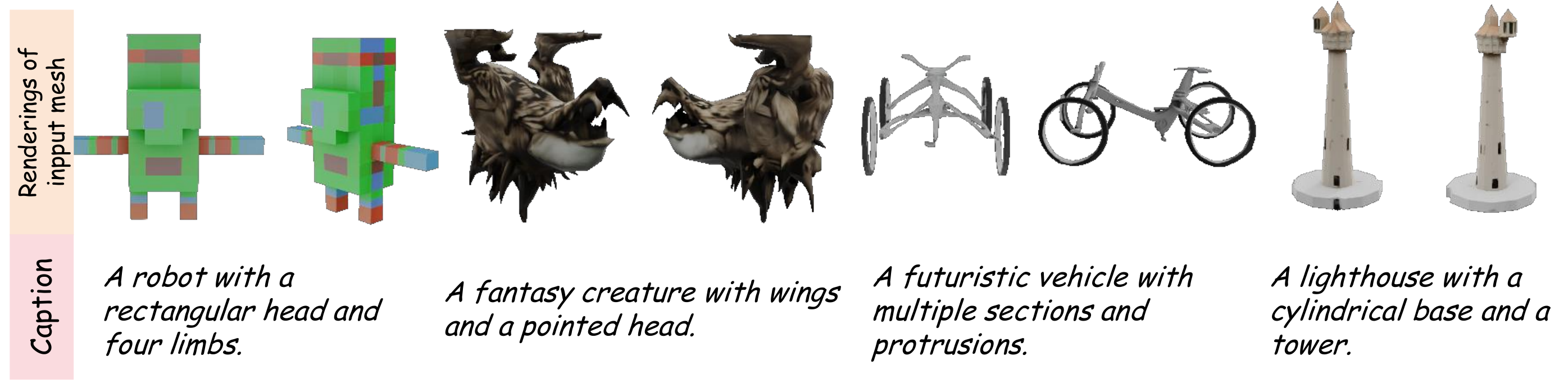}
    \vspace{-0.03\textheight}
    \caption{\textbf{Some cases of 3D editing result from our method.}}
    \label{fig:caption_result}
\end{figure}
\begin{figure}[th]
    \centering
    \includegraphics[width=\linewidth]{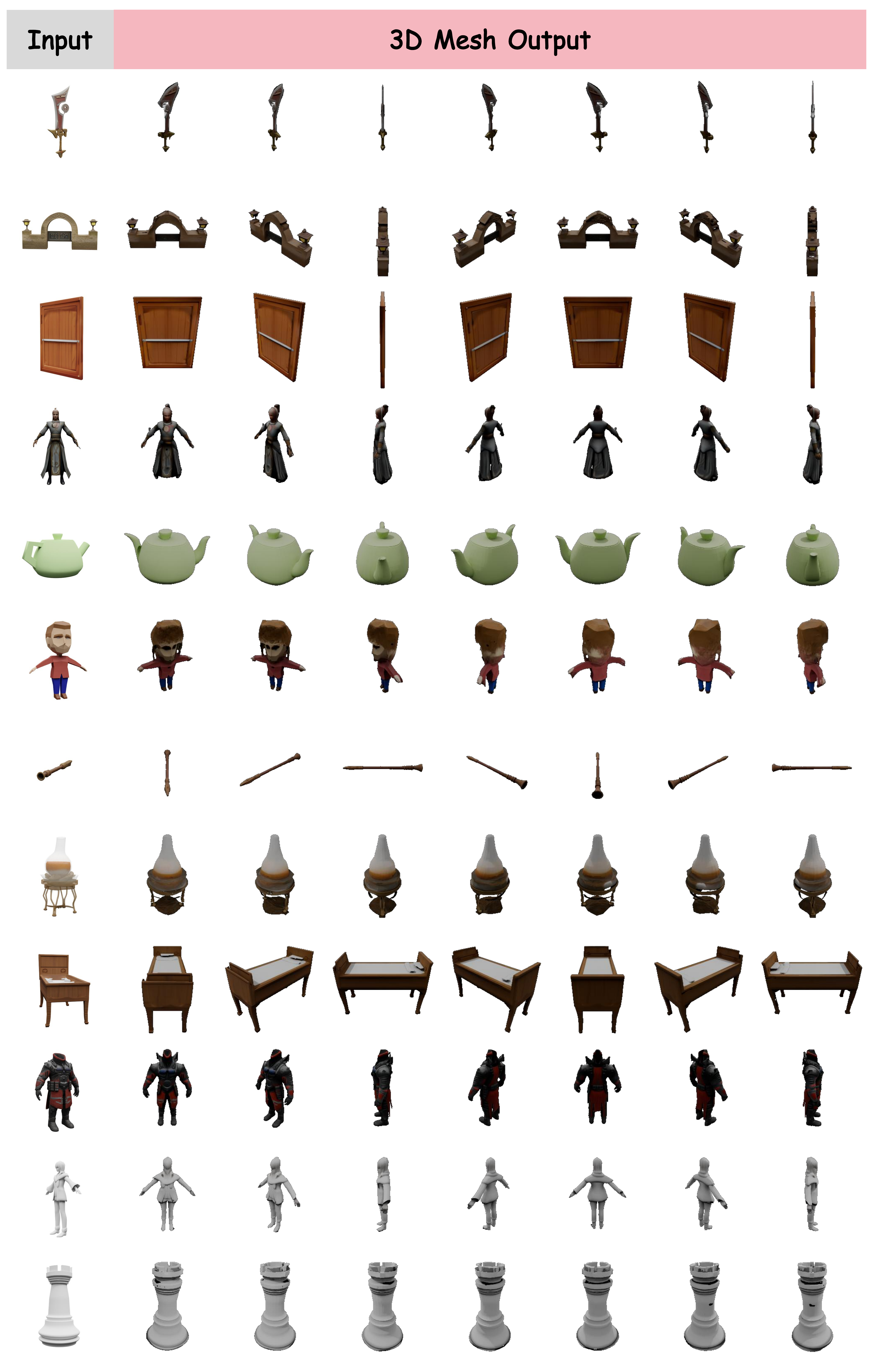}
    \caption{\textbf{More cases of Image-to-3D result from our method.}}
    \label{fig:image_to_3d_1}
\end{figure}
\begin{figure}[th]
    \centering
    \includegraphics[width=\linewidth]{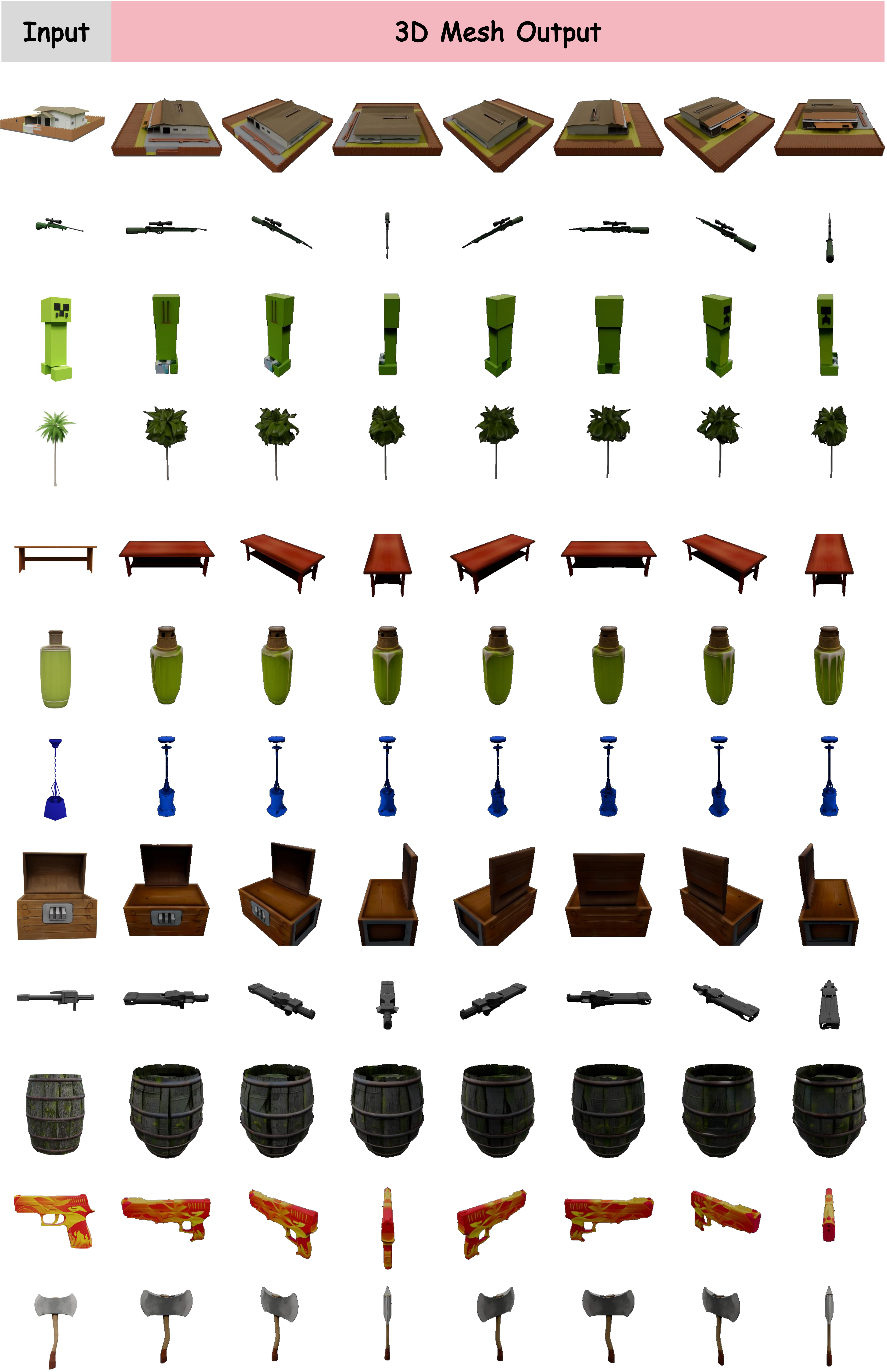}
    \vspace{-0.03\textheight}
    \caption{\textbf{More cases of Image-to-3D result from our method.}}
    \label{fig:image_to_3d_2}
\end{figure}
\begin{figure}[th]
    \centering
    \includegraphics[width=\linewidth]{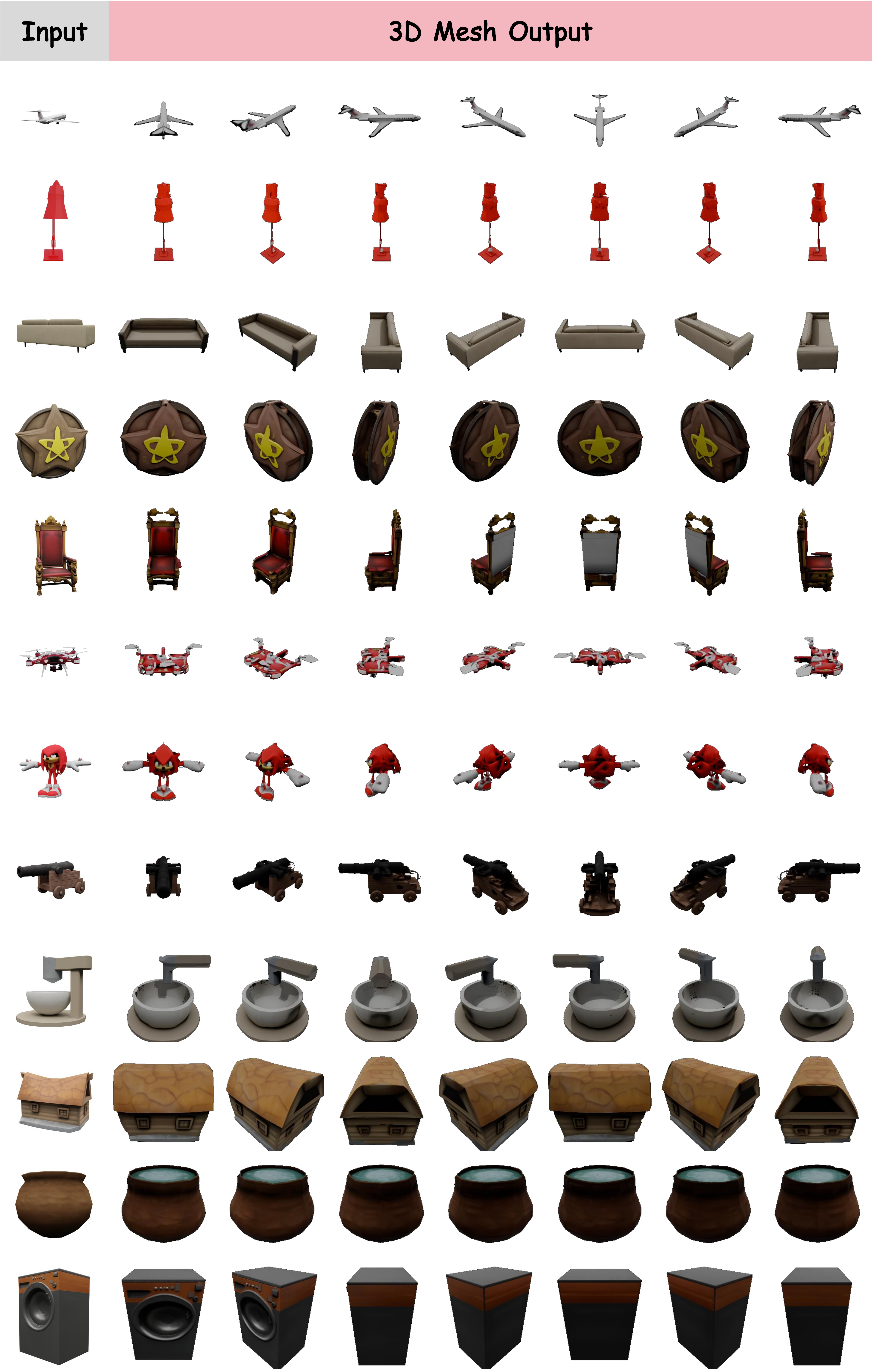}
    \vspace{-0.03\textheight}
    \caption{\textbf{More cases of Image-to-3D result from our method.}}
    \label{fig:image_to_3d_3}
\end{figure}
\begin{figure}[th]
    \centering
    \includegraphics[width=\linewidth]{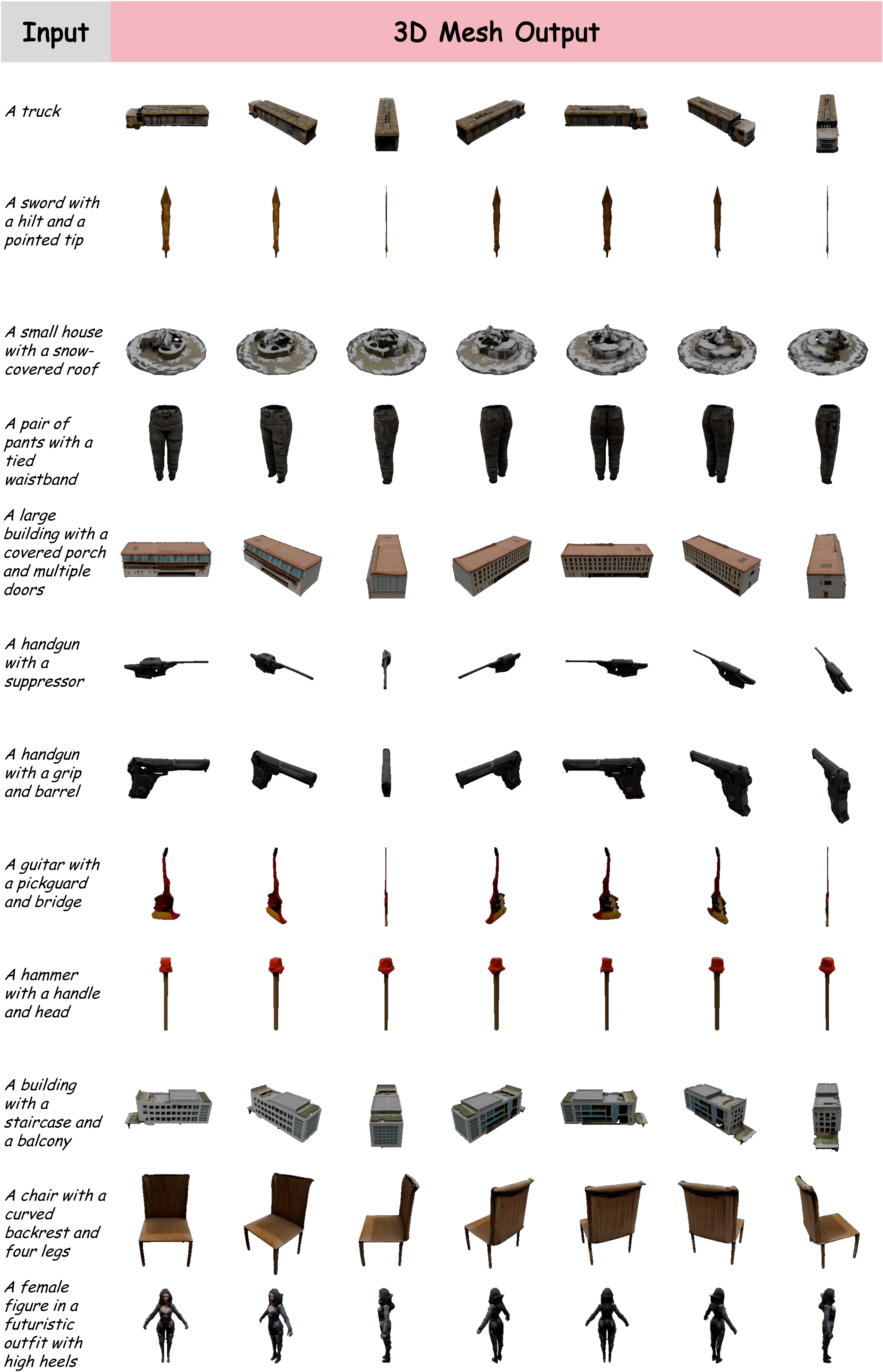}
    \vspace{-0.03\textheight}
    \caption{\textbf{More cases of Text-to-3D result from our method.}}
    \label{fig:text_to_3d_1}
\end{figure}

In Figure~\ref{fig:image_to_3d_1}, Figure~\ref{fig:image_to_3d_2}, and Figure~\ref{fig:image_to_3d_3}, we showcase additional Image-to-3D generation results. To maintain consistency with the training setup, all input images are resized to 512×512 resolution with a white background. This preprocessing step is crucial, as our base model, Qwen-VL~\cite{bai2025qwen25vl}, encodes images into token sequences whose length depends on the input resolution. Additional Text-to-3D generation examples are presented in Figure~\ref{fig:text_to_3d_1}. The visual results clearly demonstrate that our model is capable of producing high-fidelity 3D assets through a unified architecture. Furthermore, Figure~\ref{fig:caption_result} provides additional 3D-to-caption generation results, and Figure~\ref{fig:caption_result2} shows two caption examples from Objaverse~\cite{deitke2023objaverse}. The generated captions demonstrate that our ShapeLLM-Omni exhibits robust 3D understanding capabilities.
\subsection{More Quantitative comparisons}
\paragraph{Ablation Study}
To determine the optimal codebook size for our 3D VQVAE model, we train several variants with different codebook sizes. We randomly sample 1000 meshes from the test dataset, voxelize them, and encode them into discrete tokens using each model. These tokens are then decoded into voxel grids and converted back to meshes through a voxel-to-mesh decoder. We evaluate reconstruction quality using Chamfer Distance (CD) and Hausdorff Distance (HD). As shown in Table~\ref{tab:ablation-study}, larger codebooks lead to better reconstruction performance. However, the improvement levels off beyond a codebook size of 8192, indicating saturation. We therefore choose 8192 as the final codebook size to strike a balance between quality and efficiency.
\begin{table}[ht]
\centering
\caption{\textbf{Ablation study on the codebook Size of 3D VQVAE}}
\label{tab:ablation-study}
\begin{tabular}{c|cc}
\toprule
\textbf{Vocabulary Size} & \textbf{Chamfer Distance$\downarrow$} & \textbf{Hausdorff Distance$\downarrow$} \\
\midrule
4096   & 0.0102 & 0.0561 \\
8192  & \textbf{0.0094} & \textbf{0.0525} \\
16384  & 0.0095 & 0.0534 \\
\bottomrule
\end{tabular}
\end{table}

\definecolor{mygray}{gray}{.9}
\begin{table*}[ht]
\centering
\caption{Edited Prompt Collection: Part One} 
\resizebox{\textwidth}{!}{%
\begin{tabular}{lll} 

\hline
ID & Category             &  Edited prompt                                                                                                                                                                                                                                             \\ \hline \hline
\rowcolor{mygray}
1  & Car        & \begin{tabular}[c]{@{\ }p{0.8\textwidth}@{\ }} Add a cannon to the front, Open the door, Add a roof rack, Add a rear wing, Lengthen the car body, Shorten the car body, Convert into a convertible, Change wheels to square shape, Bend the roof, Add air vents on the sides, Install a spotlight on the roof, Open the hood, Install a rear-view camera \end{tabular}                \\ 
2  & Tricycle   & \begin{tabular}[c]{@{\ }p{0.8\textwidth}@{\ }} Add a wheel, Install a small trumpet \end{tabular}                                                                               \\ 
\rowcolor{mygray}
3  & Bicycle     & \begin{tabular}[c]{@{\ }p{0.8\textwidth}@{\ }}Raise the seat, Add a wheel, Install a basket\end{tabular}                                                                                                                                                                 \\ 
4  & Traffic light          & \begin{tabular}[c]{@{\ }p{0.8\textwidth}@{\ }} Add an extra light, Install a surveillance camera\end{tabular}                                                                                                                           \\ 
\rowcolor{mygray}
5  & Spaceship      & \begin{tabular}[c]{@{\ }p{0.8\textwidth}@{\ }} Add wings, Add jet flames, Add solar panels, Install radar antenna, Shorten fuselage, Bend the tail fins downward, Bend the tail fins upward, Widen the wingspan, Narrow the wingspan, Tilt the whole body, Mount small missiles on wings\end{tabular}                                                                                                                                                                                              \\
6  & Tank          & \begin{tabular}[c]{@{\ }p{0.8\textwidth}@{\ }} Rotate cannon to the side, Mount a telescope on the turret top \end{tabular}                                                                                                                                              \\ 
\rowcolor{mygray}
7  &     Character      & \begin{tabular}[c]{@{\ }p{0.8\textwidth}@{\ }} Raise both hands, Raise left hand, Raise right hand, Hold a sword, Enlarge the head, Sit cross-legged, Wear a backpack, Wear a shoulder bag, Change to running pose, Grow a pair of wings, Stand on wind-fire wheels, Step on rocket launchers, Wear glasses, Wear a tall hat, Spread arms, High knee movement, Stand on one leg, Add a cape, Hold a shield, Grow a tail, Twist the waist, Stand on a skateboard, Change hairstyle to a bun, Enlarge the ears, Bend the elbows, Wear armor, Kneel on both legs, Cross both arms, Add halo above the head\end{tabular}                                                                                                                                                                          \\ 
8  & Robot        & \begin{tabular}[c]{@{\ }p{0.8\textwidth}@{\ }} Turn feet into wheels, Turn hands into bayonets, Wear an Iron Man helmet, Lengthen the arms, Mount mechanical wings on the back, Add antenna to the head, Add springs to the soles, Mount a rocket booster on the back, Lengthen the legs, Turn hands into cannons, Turn hands into claws, Turn arms into chainsaws, Add solar panels to the back, Transform into spider legs\end{tabular}                                                                                            \\ 
\rowcolor{mygray}
9  & Table         & \begin{tabular}[c]{@{\ }p{0.8\textwidth}@{\ }} Put a vase on the table, Change table shape to round, Lay a tablecloth, Spiral-shaped table legs, Add a drawer under the tabletop, Jagged edges on the tabletop, Dig a hole in the center, Put a cup on the table, Add wheels under table legs, Put a fruit plate on the table
\end{tabular}                                                                                                                                                 \\ 
10 & Chair           & \begin{tabular}[c]{@{\ }p{0.8\textwidth}@{\ }}Place a cushion, Extend the legs, Shorten the legs, Add wheels to the feet, Install a footrest, Place a seat cushion, Add storage bags on the sides, Put a speaker on it, Turn into a rocking chair\end{tabular}                                                                                                                                                                                               \\ 
\rowcolor{mygray}
11 & Cabinet    & \begin{tabular}[c]{@{\ }p{0.8\textwidth}@{\ }} Add cabinet doors, Open the cabinet doors, Add drawers, Pull out a drawer, Put a table lamp on top, Add a lock, Add internal shelves, Place a potted plant on top
Bowl: Change to square, Put an egg inside, Add a pair of chopsticks\end{tabular}                                                                                                                                                  \\ 

12 & Bed      & \begin{tabular}[c]{@{\ }p{0.8\textwidth}@{\ }} Add a pillow, Change to round shape, Add bed curtains, Place a kitten on the bed, Convert into a bunk bed\end{tabular}                                                                                                                                         \\ 
\rowcolor{mygray}
13 & Sofa     & \begin{tabular}[c]{@{\ }p{0.8\textwidth}@{\ }} Place a blanket, Place a teddy bear, Add a throw pillow\end{tabular}                                                                                                                                              \\ 
14 & Bowl & \begin{tabular}[c]{@{\ }p{0.8\textwidth}@{\ }} Change to square, Put an egg inside, Add a pair of chopsticks\end{tabular}                                                                                                                                                          \\ 
\rowcolor{mygray}
15 & Backpack     &   \begin{tabular}[c]{@{\ }p{0.8\textwidth}@{\ }}   Transform into a jetpack, Transform into a rolling backpack\end{tabular}                                                                                                                                                                                                                                                  \\
16 & Gun       &     \begin{tabular}[c]{@{\ }p{0.8\textwidth}@{\ }}  Lengthen the barrel, Add barrels on both sides, Mount a scope on top, Add a magazine slot on the left, Attach a bayonet under the muzzle  \end{tabular}                                                                                                                                                                                                                                                  \\ 
\rowcolor{mygray}
17 & Shoes         &                                                                                                                                  \begin{tabular}[c]{@{\ }p{0.8\textwidth}@{\ }}  Extend the upper part, Thicken the sole, Attach wind-fire wheels

\end{tabular}                                                                                                                    \\ 

18 & Clothes   & \begin{tabular}[c]{@{\ }p{0.8\textwidth}@{\ }} Convert to short-sleeve, Convert to long-sleeve, Add a scarf \end{tabular}                                                                               \\ 
\rowcolor{mygray}
19  & Hat     & \begin{tabular}[c]{@{\ }p{0.8\textwidth}@{\ }}Raise the crown, Add wings to the sides, Turn the top into animal ears\end{tabular}                                                                                                                                                                 \\ 
20  & Glasses          & \begin{tabular}[c]{@{\ }p{0.8\textwidth}@{\ }} Change to round frames, Add a head strap, Remove the frames\end{tabular}                                                                                                                           \\ 
\rowcolor{mygray}
21  & Ring      & \begin{tabular}[c]{@{\ }p{0.8\textwidth}@{\ }} Add a diamond, Remove the diamond\end{tabular}                                                                                                                                                                                              \\
22  & Knife          & \begin{tabular}[c]{@{\ }p{0.8\textwidth}@{\ }} Extend the blade, Turn into "Zangetsu" from Bleach \end{tabular}                                                                                                                                              \\ 
\rowcolor{mygray}
23  &     Sword      & \begin{tabular}[c]{@{\ }p{0.8\textwidth}@{\ }} Lengthen the blade, Wrap the blade in flames, Make the blade serrated, Add a ring guard to the hilt, Embed gems in the blade\end{tabular}                                                                                                                                                                          \\ 
24  & Teapot        & \begin{tabular}[c]{@{\ }p{0.8\textwidth}@{\ }} Change the spout length, Open the lid, Turn the spout into a chainsaw, Add a heater at the bottom\end{tabular}                                                                                            \\ 
\rowcolor{mygray}
25  & Bottle         & \begin{tabular}[c]{@{\ }p{0.8\textwidth}@{\ }} Only upper half remains, Insert a rose, Pour tea into the bottle, Replace cap with cork, Tie a label around the neck
\end{tabular}                                                                                                                                                 \\ 
25 & Cup           & \begin{tabular}[c]{@{\ }p{0.8\textwidth}@{\ }}Turn into conical flask, Add a handle, Add a lid, Insert a straw, Add a cup heater\end{tabular}                                                                                                                                                                                               \\ 
\rowcolor{mygray}
26 & Cat    & \begin{tabular}[c]{@{\ }p{0.8\textwidth}@{\ }} Jumping pose, Skating on a skateboard, Add a pair of wings, Wear clothes, Wear a bow on the head\end{tabular}                                                                                                                                                  \\ 

27 & Dog      & \begin{tabular}[c]{@{\ }p{0.8\textwidth}@{\ }} Hold a bone in mouth, Add a dog leash, Wear clothes, Wear a Christmas hat\end{tabular}                                                                                                                                         \\ 
\rowcolor{mygray}
28 & Insect     & \begin{tabular}[c]{@{\ }p{0.8\textwidth}@{\ }}Remove wings, Remove antennae, Add an antenna, Add a pair of wings\end{tabular}                                                                                                                                              \\ 
29 & Fish & \begin{tabular}[c]{@{\ }p{0.8\textwidth}@{\ }} Wear goggles\end{tabular}                                                                                                                                                          \\ 
\rowcolor{mygray}
30 & Block-shaped Object     &   \begin{tabular}[c]{@{\ }p{0.8\textwidth}@{\ }}  Be stretched\end{tabular}                                                                                                                                                                                                                                                  \\
31 & Ball-shaped Object      &     \begin{tabular}[c]{@{\ }p{0.8\textwidth}@{\ }}Change to oval \end{tabular}                                                                                                                                                                                                                                                  \\

\end{tabular}
}
\label{table:edited1} \\
\end{table*}

\begin{table*}[ht]
\centering
\caption{Edited Prompt Collection: Part Two}
\resizebox{\textwidth}{!}{%
\begin{tabular}{lll} 
\hline

ID & Category             &  Edited prompt                                                                                                                                                                                                                                             \\ \hline \hline

\rowcolor{mygray} 
32 & House         &                                                                                                                                  \begin{tabular}[c]{@{\ }p{0.8\textwidth}@{\ }} Add chimney on roof, Add and open a door, Change roof to dome, Change door to arch, Add canopy on the door, Add garage on the side, Add a balcony, Add a street lamp next to house, Add a fence, Add a mailbox at entrance, Install solar panels on roof\end{tabular}   \\

33 & Tower       &     \begin{tabular}[c]{@{\ }p{0.8\textwidth}@{\ }}  Shorten height, Add flag on top, Add door at base, Add spotlight at tip, Add fence around, Add antenna on top, Add spiral staircase outside, Add window in middle, Add vines on surface, Keep only lower half, Add observation deck at top\end{tabular}                                                                                                                                                                                                                                                  \\ 
\rowcolor{mygray}
34 &      Tree    &                                                                                                                                  \begin{tabular}[c]{@{\ }p{0.8\textwidth}@{\ }} Grow two giant hands, Grow giant flowers on top, Grow stars at top, Grow two long legs, Grow large wings on sides, Butterfly perching on tree, Add a door on trunk, Hang lanterns on branches\end{tabular}                                                                                                                    \\ 

35 & Flower   & \begin{tabular}[c]{@{\ }p{0.8\textwidth}@{\ }}Add more petals, Insert into vase, Bee perching on it \end{tabular}                                                                               \\ 
\rowcolor{mygray}
36  & Fruit     & \begin{tabular}[c]{@{\ }p{0.8\textwidth}@{\ }}Put in fruit plate, Peel skin, Insert small umbrella on surface\end{tabular}                                                                                                                                                                 \\ 
37 & Vegetable          & \begin{tabular}[c]{@{\ }p{0.8\textwidth}@{\ }} Be stretched\end{tabular}                                                                                                                           \\ 
\rowcolor{mygray}
38 & Phone      & \begin{tabular}[c]{@{\ }p{0.8\textwidth}@{\ }} Turn into tri-fold screen, Add stylus on edge\end{tabular}                                                                                                                                                                                              \\
39  & Computer          & \begin{tabular}[c]{@{\ }p{0.8\textwidth}@{\ }} Grow wheels \end{tabular}                                                                                                                                              \\ 
\rowcolor{mygray}
40  &     TV      & \begin{tabular}[c]{@{\ }p{0.8\textwidth}@{\ }} Add two antennas, Install base stand\end{tabular}                                                                                                                                                                          \\ 
41  & Keyboard        & \begin{tabular}[c]{@{\ }p{0.8\textwidth}@{\ }}Change to round keycaps \end{tabular}                                                                                            \\ 
\rowcolor{mygray}
42  & Book         & \begin{tabular}[c]{@{\ }p{0.8\textwidth}@{\ }} Grow two arms and legs,  Grow wings
\end{tabular}                                                                                                                                                 \\ 
43 & Building           & \begin{tabular}[c]{@{\ }p{0.8\textwidth}@{\ }}Add arched entrance in front,  Install antenna on roof,  Add chimney on roof,  Add external staircase,  Add billboard on top,  Helicopter parked on roof,  Add fence in front,  Make building round,  Install solar panels on roof,  Add flag on roof,  Change door to revolving door,  Add a clock on wall,  Hang string lights on wall\end{tabular}                                                                                                                                                                                               \\ 
\rowcolor{mygray}
44 & Building Structure    & \begin{tabular}[c]{@{\ }p{0.8\textwidth}@{\ }} Remove one column,  Change to flat roof,  Convert to castle top,  Add cable support structure\end{tabular}                                                                                                                                                  \\ 

45 & Statue      & \begin{tabular}[c]{@{\ }p{0.8\textwidth}@{\ }} Add a pair of wings,  Wear sunglasses,  Wear headphones,  Wear a tall hat,  Add halo above,  Add fence around,  Add multiple arms,  Change head to Medusa,  Wear a flower crown,  Be wrapped in chains\end{tabular}                                                                                                                                         \\ 
\rowcolor{mygray}
46 & Lamp     & \begin{tabular}[c]{@{\ }p{0.8\textwidth}@{\ }}Change bulb to square,  Change lampshade shape,  Add more lamp heads,  Change lamp head direction,  Add hanging chains\end{tabular}                                                                                                                                              \\ 
47 & Door & \begin{tabular}[c]{@{\ }p{0.8\textwidth}@{\ }}Replace rectangle door with arch,  Add doorbell,  Add surveillance camera,  Add door lock,  Add steps at entrance,  Open the door,  Wrap door with vines,  Add peephole
Bird: Claw grasping branch,  Wings spread,  Pecking downward,  Lengthen beak,  Shorten beak,  Wear top hat,  Hold a branch in beak,  Wear goggles\end{tabular}                                                                                                                                                          \\ 
\rowcolor{mygray}
48 & Sculpture     &   \begin{tabular}[c]{@{\ }p{0.8\textwidth}@{\ }}  Wear crown,  Wear armor,  Wear mask,  Hold scepter\end{tabular}                                                                                                                                                                                                                                                  \\
49 & Weapon      &     \begin{tabular}[c]{@{\ }p{0.8\textwidth}@{\ }}Add hook at front,  Make blade wavy,  Change to double-headed,  Be chained \end{tabular}                                                                                                                                                                                                                                                  \\ 
\rowcolor{mygray}
50 & Helmet         &                                                                                                                                  \begin{tabular}[c]{@{\ }p{0.8\textwidth}@{\ }} Add goggles,  Add visor,  Change to pointed top,  Unfold side wings
\end{tabular}  \\

51 & Bridge      &     \begin{tabular}[c]{@{\ }p{0.8\textwidth}@{\ }}Convert to suspension bridge,  Add pillars,  Make multi-level,  Add street lights,  Add toy cars \end{tabular}                            \\                                                                                                                   \rowcolor{mygray} 
52 & Vase         &                                                                                                                                  \begin{tabular}[c]{@{\ }p{0.8\textwidth}@{\ }} Insert flowers,  Place on table,  Add handles on sides\end{tabular}   \\

53 & Mechanical Arm       &     \begin{tabular}[c]{@{\ }p{0.8\textwidth}@{\ }}  Replace hand with clamp,  Arm rotates\end{tabular}                                                                                                                                                                                                                                                  \\ 
\rowcolor{mygray}
54 &      Plant    &                                                                                                                                  \begin{tabular}[c]{@{\ }p{0.8\textwidth}@{\ }} Add fruits,  Broken branches,  Grow upwards
\end{tabular}                                                                                                                    \\ 

55 & Shield   & \begin{tabular}[c]{@{\ }p{0.8\textwidth}@{\ }}Change to octagonal,  Embed gem in center,  Insert an arrow,  Wrap in vines\end{tabular}                                                                               \\ 
\rowcolor{mygray}
56  & Chest     & \begin{tabular}[c]{@{\ }p{0.8\textwidth}@{\ }} Be flattened,  Open lid,  Lock with chains 
\end{tabular}                                                                                                                                                                 \\ 
57 & Airplane          & \begin{tabular}[c]{@{\ }p{0.8\textwidth}@{\ }} Mount missiles under wings,  Retract landing gear,  Extend landing gear,  Add more engines\end{tabular}                                                                                                                           \\ 
\rowcolor{mygray}
58 & Castle      & \begin{tabular}[c]{@{\ }p{0.8\textwidth}@{\ }}Add drawbridge at entrance,  Attach a dragon on wall,  Connect towers with bridges,  Hang flags on walls\end{tabular}                                                                                                                                                                                              \\
59  & Mythical Creature          & \begin{tabular}[c]{@{\ }p{0.8\textwidth}@{\ }}Add saddle,  Grow spikes on back,  Sleep curled on ground \end{tabular}                                                                                                                                              \\ 
\rowcolor{mygray}
60  &    Pillar     & \begin{tabular}[c]{@{\ }p{0.8\textwidth}@{\ }}Change to polygonal,  Bend to one side,  Add grooves to body\end{tabular}                                                                                                                                                                          \\ 
61  & Tool        & \begin{tabular}[c]{@{\ }p{0.8\textwidth}@{\ }} Lengthen handle,  Replace tool head with bayonet,  Bend the handle \end{tabular}                                                                                            \\ 
\rowcolor{mygray}
62  & Lighthouse         & \begin{tabular}[c]{@{\ }p{0.8\textwidth}@{\ }} Add radar antenna on top,  Add spiral staircase outside,  Add window
\end{tabular}                                                                                                                                                 \\ 
63 & Box           & \begin{tabular}[c]{@{\ }p{0.8\textwidth}@{\ }}Be flattened,  Open the lid,  Punch a hole\end{tabular}                                                                                                                                                                                               \\ 
\rowcolor{mygray}
64 & Monument    & \begin{tabular}[c]{@{\ }p{0.8\textwidth}@{\ }} Change top to pointed,  Add flag on top,  Add steps at base\end{tabular}                                                                                                                                                  \\ 

65 & Animal      & \begin{tabular}[c]{@{\ }p{0.8\textwidth}@{\ }} Grow antennae\end{tabular}                                                                                                                                         \\ 
\rowcolor{mygray}
66 & Stairs     & \begin{tabular}[c]{@{\ }p{0.8\textwidth}@{\ }}Add more steps,  Change to spiral stairs,  Remove handrails\end{tabular}                                                                                                                                              \\ 
67 & Tent & \begin{tabular}[c]{@{\ }p{0.8\textwidth}@{\ }}Extend awning,  Change to dome-shaped\end{tabular}                                                                                                                                                          \\ 
\rowcolor{mygray}
68& Street Light     &   \begin{tabular}[c]{@{\ }p{0.8\textwidth}@{\ }}  Add signboard on pole,  Add camera on pole\end{tabular}                                                                                                                                                                                                                                                  \\
69 & Trophy      &     \begin{tabular}[c]{@{\ }p{0.8\textwidth}@{\ }}Add lid,  Add handles \end{tabular}                                                                                                                                                                                                                                                  \\ 
\rowcolor{mygray}
70 & Machine         &                                                                                                                                  \begin{tabular}[c]{@{\ }p{0.8\textwidth}@{\ }}Add wheels\end{tabular}  \\

\end{tabular}
}
\label{table:edited2} \\
\end{table*}
\newpage
{
\small
\bibliographystyle{plainnat}
\bibliography{neurips_2025}
}

\end{document}